**Hierarchical Stacking Optimization Using Dirichlet's Process (SoDip): Towards Accelerated Design for Graft Polymerization**

*Amgad Ahmed Ali Ibrahim \*, Hein Htet, and Ryoji Asahi \*\**

Institutes of Innovation for Future Society, Nagoya University,
Furo-cho, Chikusa-ku, Nagoya, Aichi 464-8603, Japan

\*\*E-mail: asahi.ryoji
\*  E-mail: ibrahim.amgad

**Abstract**

Radiation-induced grafting (RIG) enables precise functionalization of polymer films for ion-exchange membranes, $CO_2$-separation membranes, and battery electrolytes by generating radicals on robust substrates to graft desired monomers. However, reproducibility remains limited due to unreported variability in base-film morphology (crystallinity, grain orientation, free volume), which governs monomer diffusion, radical distribution, and the Trommsdorff effect, leading to spatial graft gradients and performance inconsistencies. We present a hierarchical stacking optimization framework with a Dirichlet's Process (SoDip), a hierarchical data-driven framework integrating: (1) a decoder-only Transformer (DeepSeek-R1) to encode textual process descriptors (irradiation source, grafting type, substrate manufacturer); (2) TabNet and XGBoost for modelling multimodal feature interactions; (3) Gaussian Process Regression (GPR) with Dirichlet Process Mixture Models (DPMM) for uncertainty quantification and heteroscedasticity; and (4) Bayesian Optimization for efficient exploration of high-dimensional synthesis space. A diverse dataset was curated using ChemDataExtractor 2.0 and WebPlotDigitizer, incorporating numerical and textual variables across hundreds of RIG studies. In cross-validation, SoDip achieved ~33% improvement over GPR while providing calibrated confidence intervals that identify low-reproducibility regimes. Its stacked architecture integrates sparse textual and numerical inputs of varying quality, outperforming prior models and establishing a foundation for reproducible, morphology-aware design in graft polymerization research.

## 1. Introduction

Functional polymers are widely recognized for their distinctive properties, which are governed by the careful control of synthesis conditions [1–5]. Radiation-induced grafting (RIG) has emerged as a powerful method for modifying preformed polymer films, offering significant advantages such as preserving the mechanical integrity of membranes and facilitating the introduction of functional groups without degrading the base polymer [1,2,5–8]. RIG enables the preparation of a wide range of advanced materials, including proton exchange membranes (PEMs), alkaline anion exchange membranes (AEMs), $CO_2$ separation membranes, and polymer electrolytes for batteries [7,9].

However, the high cost of experimental and computational methods makes it challenging to determine the optimal synthesis parameters of these processes [1–3]. Specifically, two intertwined challenges hinder progress:

1. Intrinsic experimental reproducibility issues. Commercial polymer films, even when nominally identical, differ in crystallinity, grain orientation, free-volume distribution, and thickness. These morphological variations rarely reported in published grafting studies profoundly affect radical formation, monomer diffusion kinetics, and graft yield, leading to unpredictable spatial gradients in graft concentration and inconsistent membrane performance[10–14].

2. High-dimensional, multimodal parameter spaces. RIG optimization must consider numerical continuous factors (dose, dose-rate, temperature, monomer concentration), numerical discrete factors (molecular weight of monomer, size of base polymer), and categorical textual descriptors (irradiation source, grafting method, substrate manufacturer), where an integration exceeds the capacity of traditional design-of-experiments (DOE) and polynomial regression methods, which require extensive trials and often miss complex interactions [15–17].

Conventional DOE and response-surface models become impractical as dimensionality grows and cannot incorporate textual variables. Standalone machine-learning tools (e.g., Gaussian Processes for numerical data or deep neural networks for complex patterns) either ignore categorical/textual inputs or lack proper uncertainty quantification under sparse data [16,18].

To overcome these gaps, a need for a model that can handle high-dimensional, multimodal inputs[17,19–21]. Furthermore, this model can provide calibrated uncertainty estimates for each prediction, to directly flag conditions and material batches where experimental reproducibility is likely to be poor. In practice, regions of high predictive uncertainty correlate with sparse or highly variable data (mainly caused by uncharacterized film heterogeneity)[17,22]. This uncertainty quantification enables researchers to prioritize additional experiments or characterization efforts on those high-uncertainty regimes, thereby systematically closing reproducibility gaps and improving confidence in both model predictions and experimental outcomes[22–24]. Here we introduce hierarchical stacking optimization using Dirichlet's process (SoDip), an end-to-end, hybrid machine-learning pipeline explicitly designed for RIG, where the features of the model include:

1. Transformer-based encoding using large language model (DeepSeek-R1)[25] translates grafting method, irradiation source, and supplier metadata into rich numerical embeddings, thus conquering multimodal, partially textual inputs[25–27].
2. TabNet (multimodal tabular regressor), and XGBoost capture complex, nonlinear interactions among numerical, categorical, and embedding features[28–31].
3. Gaussian Process regression within a Dirichlet-Process Mixture Model (DPMM) clustering layer provides robust uncertainty quantification and handles heteroscedastic noise by partitioning the data into locally homogeneous regimes[32–35].
4. Bayesian Optimization efficiently navigates the high-dimensional parameter space, reducing the number of costly experiments needed to identify optimal conditions[18,36].

In the current study, we systematically curated data from hundreds of RIG publications using ChemDataExtractor 2.0 and WebPlotDigitizer to assemble a comprehensive multimodal dataset of numerical and textual process descriptors of RIG [37–41]. Previous efforts have applied DOE, RSM, and /or ML to optimise RIG, such as standalone GP models for permeability prediction or deep-learning classifiers for categorical parameters. These approaches typically address only a subset of variables and lack robust uncertainty quantification, which limits their reproducibility and generalizability[16,18,23,42]. In contrast, SoDip is, a framework that simultaneously integrates most of critical RIG dimensions including irradiation source, grafting method, base-film supplier, and proxy morphology descriptors within a single, end-to-end predictive pipeline.

Earlier approaches often struggle with reproducibility because they (a) ignore unstructured textual metadata, (b) lack mechanisms to partition heterogeneous data regimes, and (c) do not provide calibrated confidence intervals for predictions. SoDip addresses these limitations by integrating Transformer-derived embeddings, multimodal tabular regression, DPMM clustering, and GPR with uncertainty quantification. By explicitly capturing nonlinear, cluster-aware relationships and leveraging rich textual and tabular features, SoDip provides a robust framework for data-driven design of functional polymer membranes.

## 2. Methodology

Functional polymers, with their tailored chemical structures and adaptable properties, present a rich and versatile data source for testing of advanced machine learning and optimization models, particularly due to their diverse synthesis methods, tuneable functionalities, and wide-ranging applications in energy, healthcare, and environmental technologies. Specifically, RIG provides a distinctive subset for model training and validation: by adjusting radiation dose, monomer composition and grafting time, RIG produces high-dimensional datasets with continuous metrics such as grafting efficiency, mechanical and thermal stability and grafting rates, that challenge models to elucidate complex non-linear relationships and predict material behaviour under defined experimental conditions, thereby enabling inverse design. Nevertheless, ensuring reproducibility and accommodating scalable data generation demand rigorous uncertainty quantification during optimization and prediction. One major challenge in modelling and optimizing RIG is the reproducibility of results even under controlled conditions ensures dataset consistency, while scalability in experimental setups supports large-volume data generation, thus a proper estimation of uncertainties is required while performing optimization and predictions. Thus, by leveraging RIG data, we can rigorously test the robustness, interpretability, and adaptability of SoDip in simulating real-world material design scenarios, accelerating innovation in smart polymer technologies.

2.1 Data collection and preparation

The efficacy of the information extraction process is significantly influenced by the quality and diversity of the data sources. In this study, we employed our automated data collection methodology, as detailed in previous literature[43] , to gather pertinent articles using a custom-developed web-based platform. These articles were used to generate a dataset for training and evaluation of SoDip.

2.1.1. Adopted Tools & Techniques

To gather the data required for this study, we employed the following tools and techniques:

• Article Database: This serves as a comprehensive repository[37,43] of scientific literature, with a particular emphasis on chemistry and materials science. We primarily utilized Elsevier and the Royal Society of Chemistry (RSC) as our main sources for article retrieval.

• ChemDataExtractor2 (CDE2): incorporates a sophisticated database scraper module[38,43], which is designed to efficiently retrieve relevant articles from various databases, thereby ensuring the collection of most relevant articles to RIG that contains preparation data of grafted films. It was used also to fetch categorical information that does not exist in figures, such as materials supplier, type of grafting techniques, used solvents, etc.

• WebPlotDigitizer 4.8 software: is a software tool designed to extract numerical data from various types of plot images, such as XY charts, bar graphs, scatter plots, and others. It uses computer vision techniques to accurately digitize data points from images; in the present study we used it to extract the numeric values for the grafting condition predictors[39–41].

2.1.2. RIG Related Articles Selection

For the selection of relevant articles, we used the query: Radiation AND Grafting AND (polymerization OR "craft copolymerization" OR "ORR kinetics" OR "grafting conditions") to search for literature. The collected articles covered a wide range of publications, including journals, conference proceedings, and technical reports, published between 1996 and 2024, ensuring the inclusion of the most up-to-date research in the field. A total of 115 articles were identified. From the full-text articles, we focused on the Results & Discussion, and figures, as these sections contain the most essential information required to form a dataset.

For the present study, 70% of the articles were collected from Elsevier, 20% from RSC, 6% American chemical society (ACS), and 4% from others. However, only 40 articles were including parametric data that is harmonious enough to develop a dataset of 1382 data point.

Other articles were rejected because at least on two parameters of the main body of our data set was missing.

The RIG problem represents a high-dimensional regression challenge characterized by a heterogeneous mix of data types, including textual categorical, and numerical variables. In this study, twenty independent predictor variables were selected as RIG descriptors alongside a single continuous response variable (grafting yield $G_y$). In this article we will denote descriptors with an uppercase D and subscript with the name of the descriptor as summarised in Table 1, for instance descriptor of absorbed dose will be denoted as $D_{Dose}$.

**Table1** list of used descriptors, and its corresponding notation and type.

| Descriptor | Notation \| Value | Type |
|---|---|---|
| Grafted film | $D_{Graft}$: Ps-g-ETFE | Textual (Categorical) |
| Monomer identity | $D_{Monomer}$: styrene | Textual (Categorical) |
| Base polymer name | $D_{Base}$: ETFE | Textual (Categorical) |
| Base polymer morphology | $D_{Morphology}$: fibre | Textual (Categorical) |
| Base polymer supplier | $D_{Supplier}$: Goodfellow | Textual (Categorical) |
| Grafting Method | $D_{Method}$: pre-irradiation | Textual (Categorical) |
| Irradiation source | $D_{Source}$: gamma | Textual (Categorical) |
| Solvent identity | $D_{Solvent}$: methanol | Textual (Categorical) |
| Additive identity | $D_{Additive}$: ferrous sulfate | Textual (Categorical) |
| Base polymer size | $D_{Size}$: 50 µm | Numeric (Discrete) |
| Molecular weight of repeating unit of base polymer film | $D_{Base\_MW}$: 64.02 g/mol | Numeric (Discrete) |
| Molecular weight of monomer | $D_{Monomer\_MW}$: 94 g/mol | Numeric (Discrete) |
| Molecular weight of solvent | $D_{Solvent\_MW}$: 32.1 g/mol | Numeric (Discrete) |
| Molecular weight of additive | $D_{Additive\_MW}$: 98 g/mol | Numeric (Discrete) |
| Absorbed Dose | $D_{Dose}$: 100 kGy | Numeric (Continuous) |
| Grafting temperature | $D_{Temp.}$: 60 °C | Numeric (Continuous) |
| Grafting time | $D_{Time}$: 12 h | Numeric (Continuous) |
| Monomer concentration | $D_{Monomer\_conc.}$: 5 vol% | Numeric (Continuous) |
| Solvent concentration | $D_{solvent\_conc.}$: 15 vol% | Numeric (Continuous) |
| Additive concentration | $D_{Add\_conc.}$: 10 vol% | Numeric (Continuous) |
| Grafting Yield | $G_y$: 237% | Numeric (Continuous) |

These descriptors represent the variables most frequently reported in the RIG literature. In principle, it would be preferable to replace the categorical variable $D_{Supplier}$ with quantitative measures of base-film morphology such as pores-distribution profiles, crystallinity, or free-volume fractions because these properties fundamentally govern grafting kinetics and final grafted film performance. However, such morphological data rarely appear in grafted film studies, since they require extensive pre-grafting characterization (e.g. microinterferometry, X-ray scattering, atomic forced microscopy, etc.) that is typically reported only in separate materials science articles. Moreover, to include these parameters in our dataset, every source paper would need to provide them, which is not the case. Therefore, we have retained $D_{Supplier}$ as a proxy for all unreported morphological variability, on the assumption that supplier quality control yields films with comparable morphology.

These predictors are spatially distributed across various locations within the RIG experimental space of parameters, thereby introducing significant complexity for conventional probabilistic optimization and predictive modelling techniques. Such spatial heterogeneity and data diversity necessitate advanced modelling strategies to ensure accurate inference and prediction.

In response to these challenges, we propose our hybrid-metaheuristic model that we named SoDip as depicted in Figure 1. SoDip is a multi-step regression framework designed to systematically address the intricacies of this high-dimensional, heterogeneous regression problem. The subsequent sections will portray the stages of this model, detailing the methodologies employed to accommodate the diverse data types and spatial dependencies inherent in the RIG problem. Data were provided into different types of embedding as depicted in Figure1(a) by textual, and numeric input data. Each row of the formulation metrics data is converted into a space-separated strings. Then, each textual descriptor label is concatenated with the corresponding formulation metric string and stored into an array of strings, where a single row of categorical input data might be represented as "PS-g-ETFE_ Styrene_ ETFE_ Film_ Goodfellow_ 125_ pre-iradiation_ Electorn-Beam_ methanol_ 0.4_ None_0". These categorical input data are later fed into the deepseek transformer model.

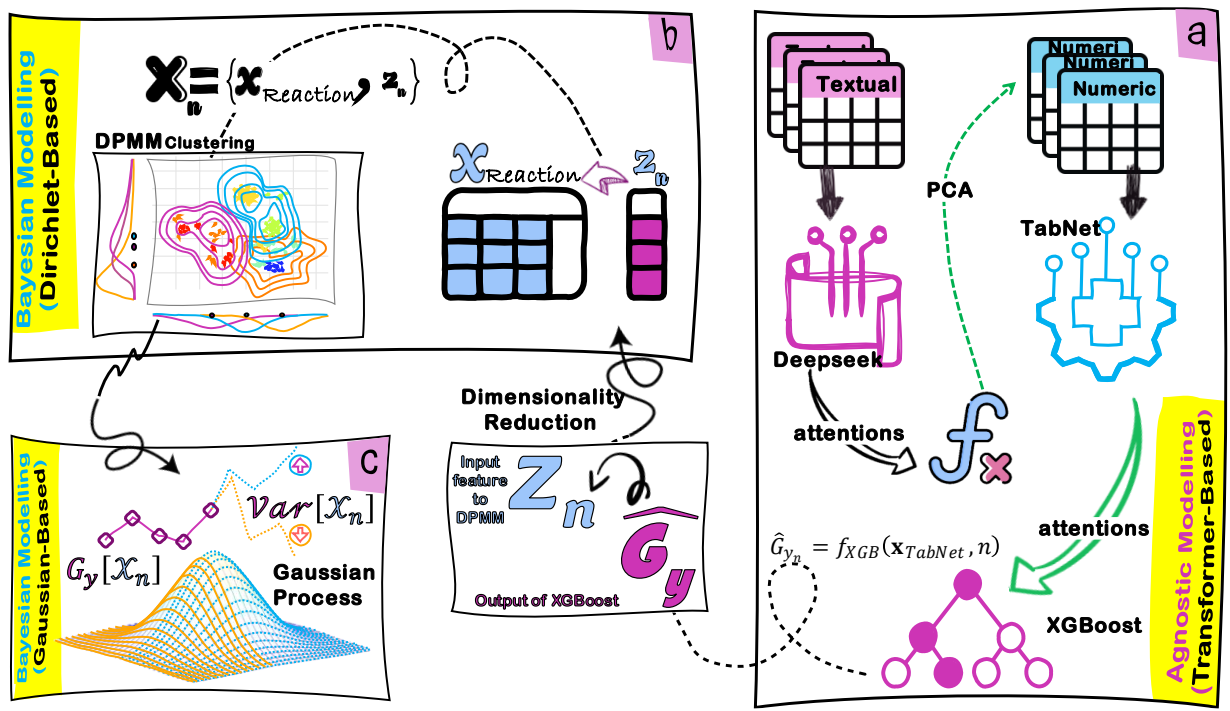

**Figure 1.** Schematic representation of main building blocks of SoDip model namely (a) agnostic learning dimensionality reduction, (b) DPMM Clustering, and (c) Gaussian process regression.

2.2 SoDip Architecture

SoDip provides a set of features as it merges categorical and numeric features into a text format for transformer processing, while maintaining regression targets. As can be seen in Figure1, SoDip is composed of three main stages explicitly (a) agnostic learning dimensionality reduction, (b) DPMM Clustering, and (c) Gaussian process regression. It creates an end-to-end regression pipeline that uses transformer-based models to process embeddings, applies an attention mechanism, and appends regression data to these embeddings.

Initially, meaningful features are extracted by DeepSeek transformer. Context-rich embeddings from DeepSeek-R1 (attention-processed hidden states) concatenated with the numeric predictors (per sample), scaled, and PCA-reduced then trains TabNet (a multimodal regressor) together with regression labels (the actual grafting yields) as presented in Figure1(a). SoDip utilizes BO for hyperparameter optimization of a XGBoost regressor that ultimately predicts the regression target after being trained on attention-processed hidden states output from TabNet, in other words, SoDip uses TabNet as a feature generator for training XGBoost. Cross-validation is used to ensure proper fitting throughout all the stages of the model.

The first stage of SoDip is dimensionality reduction, wherein all categorical textual descriptors are reduced into a single vector of unified intermediate variable through agnostic learning–based regression. This intermediate variable is subsequently incorporated together with the grafting condition numeric predictors into a suite of probabilistic Bayesian clustering and regression blocks, specifically DPMM and GPR as represented in Figure 1(b), and Figure 1(c) respectively. Such architecture where models are built on outcomes of previous models' outputs within same framework, make SoDip to be a model-of-models and hence, our approach could be called a metamodeling approach.

2.2.1 Agnostic-learning and dimensionality reduction

DeepSeek-R1 serves as the backbone for processing of unstructured text data, it converts the combined text inputs (which merge textual and numeric composition data) into high-dimensional embeddings that capture deep semantic and contextual information. This process turns raw textual data into a structured representation that can be further processed.

In SoDip the hidden states are extracted from DeepSeek-R1 transformer's final layer and are then used to compute an attention-based representation of inputs. By calculating the scaled dot-product and applying a softmax (represented by $f_x$ in Figure 1(a)), it derives attention weights that highlight relationships within the token sequence. This step enhances the original embeddings with context-aware features, which are crucial for capturing subtle patterns in the data[44,45].

The enriched embeddings from DeepSeek-R1 are then concatenated with the numeric continuous data and reshaped. These enriched features are the input to the subsequent PCA and TabNet processing. In this way, DeepSeek-R1 lays the groundwork by transforming raw text into a robust, high-level feature space that subsequent models can exploit.

The softmax here is applied as part of the attention mechanism operating on the transformer's hidden states. It converts raw attention scores (computed via the scaled dot product) into a probability distribution. This means that for each token in the sequence, the weights sum to $1$[44,45]. This normalization helps the model focus on the most relevant (towards regression targets) parts of the input by assigning higher weights to more important tokens while diminishing less important ones.

DeepSeek-R1 is designed for handling unstructured data (like text) and capturing contextual relationships. TabNet, on the other hand, is optimized for tabular data. It uses sequential attention to perform instance-wise feature selection, effectively identifying and weighting the most relevant features for each prediction. By processing the PCA-reduced outputs from DeepSeek, TabNet can capture non-linear interactions and dependencies that might not be evident in the original transformer embeddings. This refined representation can improve the predictive power of the final XGBoost model. TabNet inherently offers interpretable attention masks that highlight which features (or parts of the input) are being focused on during prediction. This can provide insights into the model's decision-making process, an advantage over the "black-box" nature of many transformer models. The combination leverages the strengths of both architectures. While DeepSeek-R1 captures rich linguistic and contextual nuances from text, TabNet focuses on learning complex relationships within the structured (tabular) feature space, leading to a more robust end-to-end pipeline.

The first stage of SoDip focuses on dimensionality reduction, wherein 16 descriptors (categorical textual, and numerical discrete) are reduced into a single unified intermediate variable through agnostic learning–based regression. The dimensionality reduction technique used in SoDip is developed from the stacked regressions approach (also known as hierarchical stacking generalization), where it was reported that feeding a cross validated first stage predictor into a second stage learner is a valid approach for dimensionality reduction and almost outperform the single best possible model. Besides, it simply preconditions the meta-feature for better downstream learning[46–48].

In SoDip case the first-stage meta-feature is grafting yield in a sample data n as,

$$\hat{G}_{y_n} = f_{XGB}(\mathbf{x}_{TabNet}, n),$$

Where $\mathbf{x}_{TabNet}$ is the DeepSeek processed attention output from TabNet.
Since in stacking regression modelling, the output of one model becomes an input to another. Applying a monotonic, and invertible transform to the meta-feature ($\hat{G}_{y_n}$) is simply a recommended preprocessing procedure as it preserves the rank and remove any feature information redundancy while preserving information of first-stage descriptors. Since GPR's kernels implicitly assume roughly Gaussian, homoscedastic inputs, the Yeo–Johnson transform[49] as a monotonic, and invertible power transform is used. Moreover, using molecular weights as fixed weights for $\hat{G}_{y_n}$ embeds experimental insight directly into the feature space. In fact, they act like a prior on the feature scale while not bias the GPR's training process. Accordingly, in SoDip $\hat{G}_{y_n}$ is then weighted and transformed using Yeo-Johnson transform leading to the intermediate variable $\mathcal{Z}_n$ represented as

$$\mathcal{Z}_n = \psi^{YJ}(\lambda, \boldsymbol{w}\hat{G}_{y_n}) = \begin{cases} \dfrac{(\boldsymbol{w}\hat{G}_{y_n} + 1)^{\lambda} - 1}{\lambda}, & \boldsymbol{w}\hat{G}_{y_n} \geq 0, \quad \lambda \neq 0, \\ \log(\boldsymbol{w}\hat{G}_{y_n} + 1), & \boldsymbol{w}\hat{G}_{y_n} \geq 0, \quad \lambda \neq 0, \\ -\dfrac{(-\boldsymbol{w}\hat{G}_{y_n} + 1)^{2-\lambda} - 1}{2 - \lambda}, & \boldsymbol{w}\hat{G}_{y_n} < 0, \quad \lambda \neq 2, \\ -\log(-\boldsymbol{w}\hat{G}_{y_n} + 1), & \boldsymbol{w}\hat{G}_{y_n} < 0, \quad \lambda = 2, \end{cases}$$

where $\boldsymbol{w}$ is a composite molecular-weight coefficient derived from the molecular weights of the repeating units of the base polymer ($\boldsymbol{w}_f$), monomer ($\boldsymbol{w}_m$), solvent ($\boldsymbol{w}_s$), and additives ($\boldsymbol{w}_a$). In practice, $\boldsymbol{w}$ is a normalized molecular-weight factor that embeds physicochemical information directly into the feature space while preserving the scalar form of $\hat{G}_{y_n}$. Fitting the Yeo–Johnson transform parameter $\lambda$ via maximum likelihood, ensures that the transformed intermediate variable ($\mathcal{Z}_n$) approximately Gaussian. To avoid any possibility of overfitting during stacked regression, the parameter $\lambda$ was fitted to the entire dataset through a cross validation this, embed the power transform fit inside the CV loop (i.e. learn $\lambda$ on each training fold only). As long as $\lambda$ is learned without peeking at test fold outputs, there should be no sort of leaking information which satisfy the same level-one CV requirement Wolpert insists on to avoid overfitting.

### 2.2.2 Probabilistic modelling

(a) Dirichlet's process mixture model (DPMM)

RIG experiments belong to multiple spaces (e.g., irradiation source, grafting type, morphology of the base polymer, different reaction pathways, etc). common practice to model such multiple space problem is by using clustering models. Traditional Gaussian mixture models (GMMs) require predefining cluster counts[18,42,50]. Here, Dirichlet's process mixture with Gibbs sampling over Normal-Inverse-Wishart distributions (NIW) could be a very useful computational approach. DPMM group experiments into clusters using Bayesian non-parametric methods, where Gibbs sampling Iteratively assign data points to clusters and update parameters, while cluster means and covariances are modelled with NIW distributions to handle uncertainty[34,35,51].

Gibbs sampling, a Markov chain Monte Carlo (MCMC) strategy, repeatedly resamples each variable from its conditional distribution given the current values of all other variables, thereby constructing a Markov chain that converges to the desired posterior[50,52]. In mixture modelling, the central variables are the class labels $C_i$. Their conditional probability is given by[33,53]

$$p(C_i = k|C_{-i}, X) \propto p(X|C)p(C_i = k|C_{-i})$$

where $p(C_i = k|C_{-i}, X)$ is the prior probability of assigning data point $i$ to class $k$, based on all other assignments, and $p(X|C)$ is the data likelihood under the full assignment[33,53,54].

For finite mixtures with a Dirichlet prior on the mixture weights, marginalizing out those weights resulting in

$$p(C_i = k|C_{-i}) = \frac{m_{-i,k} + \frac{\alpha}{k}}{n - 1 + \alpha}$$

where $m_{-i,k}$ is the count of observations in class $k$ excluding the $i^{th}$. In the infinite mixture limit embodied by the Chinese Restaurant Process (CRP) this conditional becomes

$$p(C_i = k|C_{-i}) = \begin{cases} \frac{m_{-i,k}}{n - 1 + \alpha}, & for\ new\ data\ enters\ existing\ cluster, m_{-i,k} > 0 \\ \frac{\alpha}{n - 1 + \alpha}, & for\ new\ data\ enters\ new\ cluster, k = K_{-i,+} + 1 \\ 0, & Otherwise \end{cases}$$

In such way, DPMM adapts to indefinite data structures by discovering and defining subgroups in RIG space[33,53]. It provides probabilistic cluster assignments for robust downstream modelling. Consequently, cluster assignments guide cluster-specific GPR training and predictions, while posterior probabilities (DPMM posterior of test data) are used to weight GPR predictions.

(b) Gaussian Process Regression (GPR)

While the DPMM clustering is performed probabilistically, different clusters may represent different RIG processes with specific kinetics and grafting behaviour, for instance one cluster may represent grafted copolymers produced by EB irradiation of fluorinated copolymers, while another cluster is assigned for grafted films produced by $\gamma$ -rays bombardment of polyaryletherketone. This may suggest distinct grafting reaction conditions (e.g., grafting temperature, Absorbed dose, monomer concentration, etc) per cluster. Thus, the final stage of the SoDip involves cluster-specific modelling, where training a separate GPR model for each DPMM cluster takes place.

The GPR can efficiently handles non-linear relationships (e.g., between grafting temperature and grafting yield). Besides, it enabled the use of optimizable kernel that could be tailored based on radial basis functions, Matern, and squared exponential kernels[18,55] are used to capture local correlations within clusters as in[42,56–58]

For each DPMM cluster $C = i$, the training feature vector:

$$x_n^{(i)} = x_n|(C = i) = [Z_{n,i}, D_{n,Dose,i}, D_{n,Temp,i}, D_{n,Time,i}, D_{n,Monomer\_Conc,i}] \in \mathbb{R}^5,$$

The cluster-specific Gaussian Process prior over grafting yield is expressed as a distribution over latent function: $G_y(\cdot| C = i) \sim \mathcal{GP}(0, k^{(i)}(\cdot,\cdot))$,

and the corresponding training targets for the $i^{th}$ cluster follow a multivariate normal:

$$\boldsymbol{G_y}|X^{(i)}, (C = i) \sim \mathcal{N}\left(0, K_c^{(i)}(x, x') + \sigma^2 \delta_{xx'}\right),$$

Cluster-specific predictive mean: $\mu_*^{(i)} \equiv \mathbb{E}\left[G_{y_*}|x_*, X, \boldsymbol{G_y}, (C_* = i)\right] = k_*^\top\left(K_c^{(i)} + \sigma^2 I_N\right)^{-1} \boldsymbol{G_y}$,

In addition to providing posterior mean prediction, the GPR yields expression for the predictive covariance of grafting yields quantitatively as[56–58]:

$$\sum{}_*^{(i)} \equiv Var\left[G_{y_*}|x_*, X, \boldsymbol{G_y}, (C_* = i)\right] = k^{(i)}(x_*, x_*) - k_*^\top\left(K_c^{(i)} + \sigma^2 I_N\right)^{-1} k_*,$$

where,

$k_* = [k^{(i)}(x_*, x_1), \ldots k^{(i)}(x_*, x_N)]^\top$,

$K_c^{(i)}$ is the $N \times N$ Gram (kernel) matrix with $k^{(i)}(x_n, x_m)$ entries and $\sigma^2 I_N$ accounts for observation noise[57,58].

In other words, in the SoDip, GPR uses DPMM tailored clusters posterior probabilities to fit models for local data patterns. Such design provides smooth predictions with uncertainty bounds estimates, needed for experimental design. Effect of DPMM parameters on GPR predictions is presented in supplementary information Figure S2 (A-C).

### 3. Results and discussion

In this section, evaluation of the performance of each modelling block within SoDip is presented, with a particular focus on the contribution of each block to the enhancement of the final prediction quality. The parameter space of RIG will be displayed over the response surface, and an analysis of RIG insights is conducted based on the outcomes produced by SoDip. Additionally, the generated subspaces of RIG will be thoroughly investigated.

#### 3.1. Analysis of SoDip response

In this section a discussion about performance of each block of SoDip. The fluctuation of performance between SoDip's blocks is also explained. The overall performance of the entire model is correspondingly evaluated.

#### 3.1.1. Transformers-based agnostic modelling

The relationship between the true values of the intermediate variable $\mathcal{Z}_n$ and its predicted value using agnostic-learning block of SoDip is presented in Figure 2. The individual direct response of isolated components of agnostic-learning blocks namely DeepSeek-R1, TabNet, and XGBoost is presented at the supported information (Figure S1) for sake of comparison and to visualize how the current architecture of agnostic-learning block had outperformed individual components.

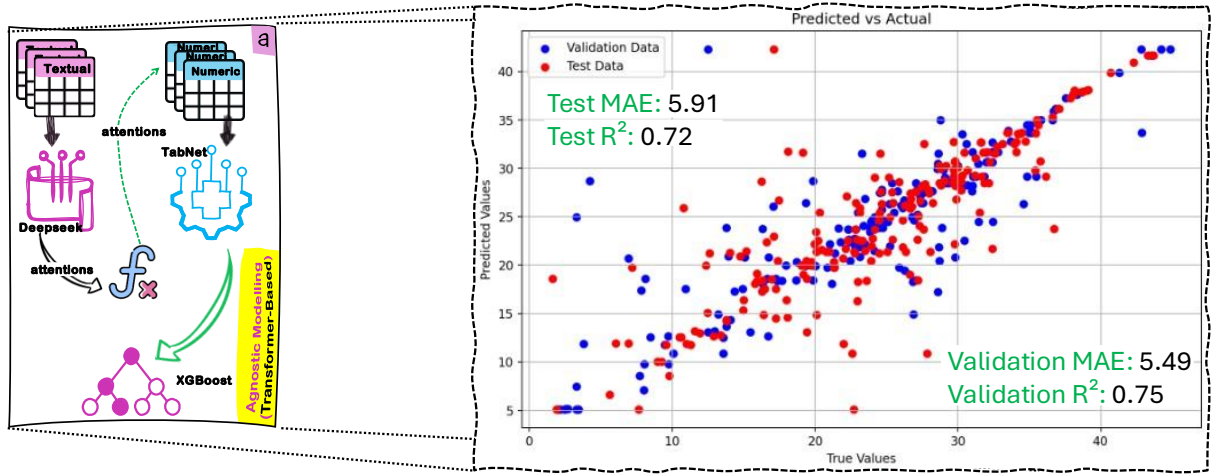

**Figure 2.** Response vs. prediction for Agnostic learning block of SoDip

Figure 2 presents a comparison of the model's predicted versus true values of $\mathcal{Z}_n$ on both the validation (blue) and test (red) sets. On the validation split, the model achieved a mean absolute error (MAE) of approximately 5.49 units and an R² of 0.75, indicating that nearly 75 % of the variance in the held-out data is captured by the predictions. Performance on the held-out test data was similarly strong, with an MAE of 5.91 units and an R² of 0.72. In both cases, the bulk of the scatter points lie tightly around the identity line (y = x), and only few observations deviated substantially from perfect prediction.

The high R² values in conjunction with low MAEs demonstrate that the model generalizes well beyond its training data, with only a modest decrease in accuracy on the test set suggesting minimal overfitting. The majority of prediction errors are concentrated at the extremes of the target variable $\mathcal{Z}_n$, which may be attributed to increased measurement noise or intrinsic heterogeneity in those regions [59–61]. This justifies the need for integrating additional domain-specific features or ensembling multiple modelling approaches could help tighten the prediction intervals and mitigate the remaining outliers [62–64]. Thus, we were motivated to implement stacked-generalization and Bayesian optimization to lift the performance as depicted in Figure 3.

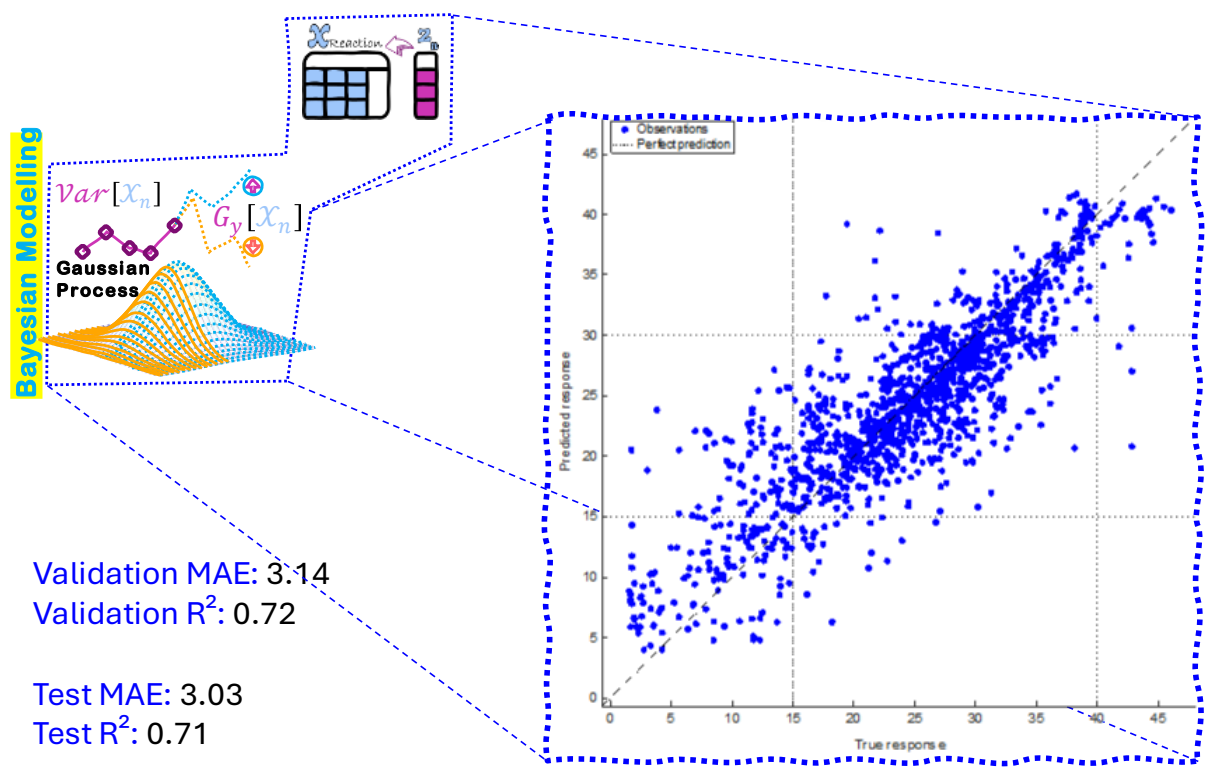

Figure 3. Illustration of the SoDip performance when only GPR is implemented directly to the outcome of staked generalization stage without clustering.

**3.1.2. Bayesian modelling and staked generalization**

Figure 3 represents the performance when GPR is implemented directly to the outcome of staked generalization stage. Specifically, it displays the cross-validation (CV) performance of the GPR model block of SoDip, plotting the true versus predicted values aggregated across all folds. The model achieved a MAE of 3.14 units and an $R^2$ of 0.72, indicating strong agreement between predictions and observations within the training regime. In particular, $R^2$ value suggests that 72 % of the variance in the validation data is explained by the model. CV was employed within the training data to optimize hyperparameters and assess model consistency, So, the CV metrics (MAE = 4.53, $R^2$ = 0.96) show how well the model generalizes during training, using different partitions of the training data. While a separate hold-out test set (untouched during model training or tuning) was used for the fair assessment of final generalization.

Performance on the held-out test set remained comparably strong, with an MAE of 3.03 units and an $R^2$ of 0.71. It is evident that, most of observations cluster closely around the dashed identity line shown on the parity plot, although a modest number of points at both the low- and high-end of the response range fall further from perfect prediction. It is important to note that the MAE and R² reported in Figure 3 derived from GPR using numeric continuous descriptors combined with the meta-variable $\mathcal{Z}_n$ cannot be directly compared to the MAE and R² in Figure 2, which reflects transformer regression predictions based on textual descriptors and numeric discrete descriptors. This improvement does not stem from a larger input space rather, it reflects the fact that the GPR operates on a more structured and continuous representation the meta-variable $\mathcal{Z}_n$ which already encapsulates information from the richer, heterogeneous descriptor space used in Figure 2. In other words, GPR benefits from a smoother, lower-dimensional feature domain, enabling more efficient learning than when trained directly on the raw textual and discrete descriptors.

### 3.1.3. Bayesian DPMM-based modelling

The cross-validated results demonstrate that the GPR effectively captures the underlying structure of the data, as evidenced by the tight clustering of points around the identity line and high R². The modest degradation in accuracy on the hold-out test set (MAE increase of 0.11 units and R² drop of 0.02) indicates that the model generalizes well, with only limited chance of overfitting. Notably, prediction errors remain small across most of the response range but grow slightly at the extremes, suggesting increased uncertainty where data are sparse. This suggested that for further improved robustness, there is a need to explore variance-stabilizing transformations or heteroscedastic regression techniques [59–61]. Additionally, integrating complementary covariates, such as domain-specific process indicators could help tighten residual variance and reduce extreme mispredictions [59,61,62]. This was achieved by integrating a DPMM for clustering, such refinements led to tighten prediction intervals and reduce the remaining outliers as presented in Figure 3.

Figure 4 illustrates the overall predictions SoDip including the effect of the integration of DPMM clustering. Cross-validation results for true versus predicted $G_y$ were plotted. Over all clusters, the cross-validation MAE is 4.53 units with an of the R² of 0.96, indicating that the model explains 96 % of the variance within the clustered folds. On the held-out test set, performance remains robust, with an MAE of 5.83 units and an $R^2$ of 0.93. Across the full $G_y$ range (0 – 450%), the bulk of the observations lie tightly about the identity line, while only a very few points of extreme values exhibit larger deviations.

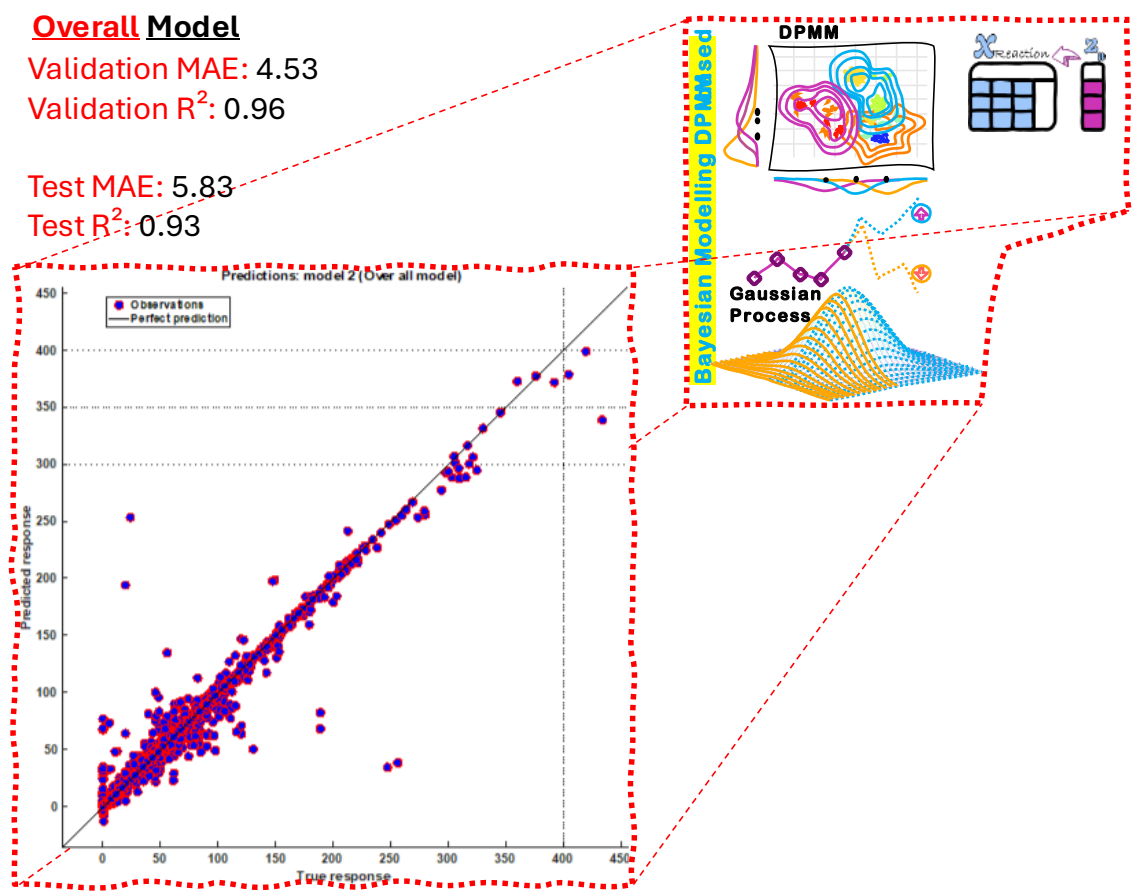

**Figure 4.** Illustration of the overall predictions SoDip including the effect of the integration of DPMM clustering. Cross-validation results for true versus predicted $G_y$ were plotted.

By first clustering the dataset via DPMM and then fitting a GPR model within each cluster, the combined framework captures local process heterogeneities more effectively than a single, global regressor. The high cross-validation $R^2$ and low MAE demonstrate excellent in-fold fit, and the modest increase in test-set error (<1.5 as a percentage of the $G_y$ range) indicates strong generalization with very limited chances of overfitting. The slight uptick in residual spread at the highest response values is likely due to sparser training data in those clusters. DPMM increased the $R^2$ value from 0.72 (in figure 3) to 0.96 (in figure 4), meaning the model explained 24 percentage points more of the data's variation. This is a relative improvement of about 33% in how much of the meaningful variation the model captures compared to GPR without DPMM. At the same time, the unexplained variance dropped from 28% to just 4%, which is a reduction of about 85% showing that the model now captures significantly greater proportion of the underlying data structure and variability. This demonstrates that the DPMM markedly strengthened the SoDip's ability to reproduce the underlying data patterns. Although the MAE in Figure 4 is marginally higher than in Figure 3, this small increase reflects the added local variability introduced by clustering. Because DPMM partitions the data into distinct regimes, each local GPR captures region-specific trends, slightly widening absolute deviations but substantially improving overall explanatory power and structural fidelity[65–68].

**3.1.4. Evaluation of SoDip's global performance and its Impact on Fabrication Reproducibility**

Reproducibility problem in RIG is inherently difficult because outcomes depend on many parameters, some of which cannot be controlled directly, such as the microscopic morphology of ready-made polymer films. As a result, repeating the same fabrication conditions does not always yield the same $G_y$s, and large variations can occur even under identical experimental conditions [1,12,13].

As shown in Sections 3.1.1–3.1.3, SoDip's modelling blocks share similar mean absolute errors (MAE) but differ markedly in coefficients of determination ($R^2$). This is expected since MAE reflects the average prediction error in physical units and is robust to outliers, while $R^2$ measures explained variance and is highly sensitive to extreme deviations. Relying on a single performance metric therefore risks obscuring important aspects of model behaviour. A balanced assessment requires multiple metrics (e.g., MAE, $R^2$), complemented by clustering analyses and prediction interval (PI) plots.

Figure 5 illustrates this evaluation. The central panel shows global cross-validated performance on the training set, while the top panel presents test-set predictions for representative clusters, and the bottom histograms display cluster-level training distributions. Using DPMM, the dataset is partitioned into nine balanced clusters, each representing most relevant fabrication regimes. SoDip achieves strong calibration, with 95% of observed points falling within the 95% PI. Detailed cluster statistics (sizes ranging from 4.8% of training data, *n*=50, to 17% of training data, *n*=187) are provided in Supplementary Table S1.

At the cluster level, SoDip captures outcome heterogeneity. Cluster 1 (n=120, R²=0.87, RMSE=16.95) displays a $G_y$ range of 0 to 450%, with bimodal tendency where one group concentrated around lower $G_y$ values 0–100%, another around 200–250%, reflecting partial stability of variance per cluster predictions as shown in cluster 1 top panel where variance increase suddenly at the mid-range of response. On the other hand, Cluster 6 (n=124, R²=0.95, RMSE=11.86) shows a unimodal, peak in the very low range 20–40 $G_y$, it is also strongly right-skew with long tail towards higher response values where about 50% of the cluster points clustered equally at the tail. Such balanced distribution explains the stable and reproducible outcomes presented at corresponding top panel for cluster 6. Cluster 9 (n=91, R²=0.91, RMSE=20.36) highlights the difficulty of reproducing many yield regimes, where distributions are right-skewed and heavy-tailed despite strong fits. Such unbalanced distribution imposed high variance peaks at many locations on the prediction per cluster as shown on top pane of cluster 9. Other clusters show similarly distinct patterns: for example, Cluster 8 is exceptionally stable (R²=0.98, RMSE=8.44, <170 $G_y$), while Clusters 3, 5, and 7 perform poorly (R²=0.12–0.36, RMSE > 17), marking unstable, irreproducible operating zones.

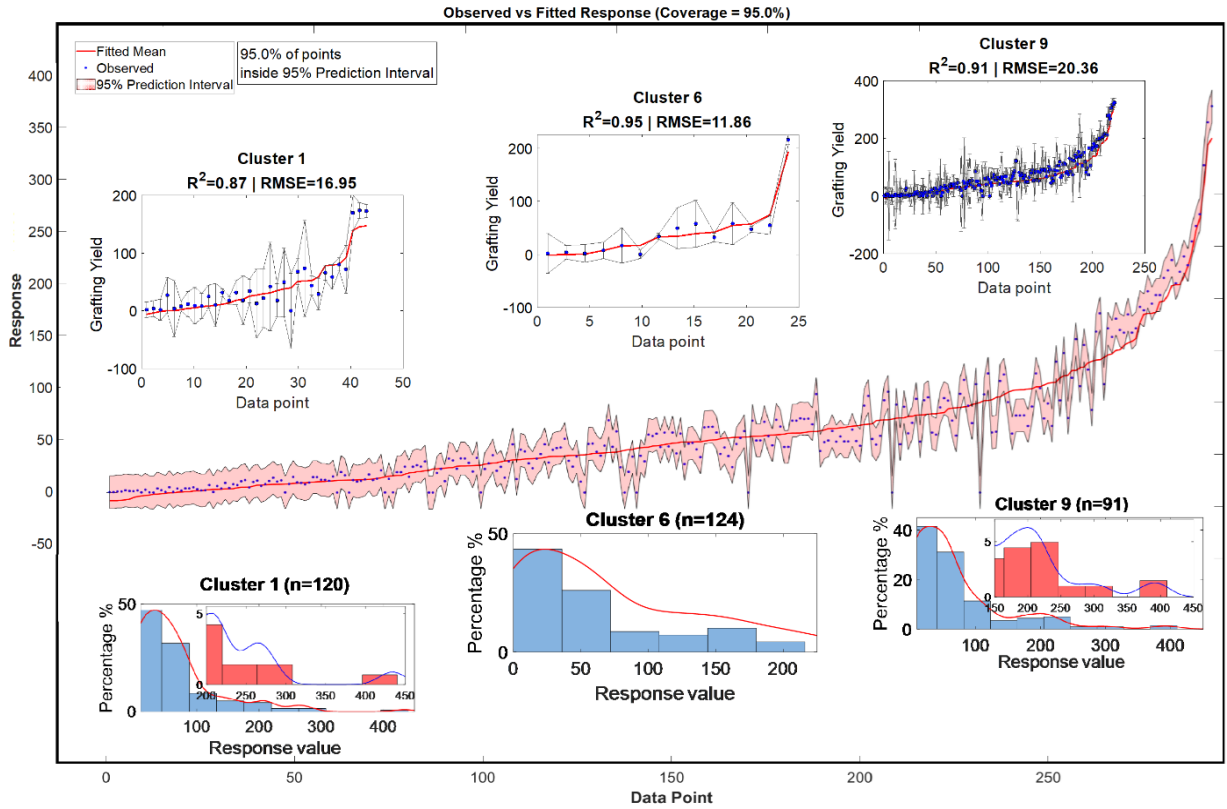

Figure 5. SoDip uncertainty of predictions plotted over observed vs predicted responses.

The link between distributions (bottom) and predictive behaviour (top/middle) is clear. In Cluster 1, frequent but widely spread $G_y$ values around 150–200% lead to broad intervals of ±60 $G_y$ in the test predictions. By contrast, the same range under the global CV fit is narrowed to less than ±25 $G_y$, since information is pooled from more stable clusters. Similarly, narrow unimodal distributions (e.g., Cluster 6) yield tight intervals (less than ±25 $G_y$), whereas broad or heavy-tailed distributions (e.g., Clusters 5, 7, 9) inflate intervals beyond ±50 $G_y$. Thus, the DPMM produces irregular but realistic uncertainty bands, widening intervals in unstable regions (Clusters 1, 9) and narrowing them in stable ones (Clusters 6, 8).

Table 2 highlights representative cases: Cluster 6 is highly stable ($R^2$=0.95, RMSE=11.86), while Cluster 9 illustrates the inherent instability of high-yield regimes (RMSE=20.36, heavy-tailed distribution).

Table 2 Cluster-Level Model Performance and Reproducibility Characteristics

| Cluster | n | R² | RMSE | Response Range ($G_y$) | Distribution Shape | Reproducibility Assessment |
|---|---|---|---|---|---|---|
| 1 | 120 | 0.87 | 16.95 | 0~450 (main: 10~100) | Bimodal, slightly right-skewed | <u>Partially stable</u> usable for low-to-moderate yields but with moderate variability. |
| 6 | 124 | 0.95 | 11.86 | 0~220 (main: 20~40) | Unimodal, highly skewed to the right | <u>Stable</u> Highly reproducible regime with low error and tight distribution. |
| 9 | 91 | 0.91 | 20.36 | 0~450 (main: 0~150) | Right-skewed, weak secondary peak | <u>Unstable</u> predicts high yield values with strong predictive fit but large variance dominates thus, reproducibility is difficult. |

Thus, SoDip provides a dual benefit: (a) Cluster-level insight – preserving local variability and identifying unstable, irreproducible conditions. (b) Global stability – borrowing strength across clusters to smooth fluctuations and tighten uncertainty where data support is strong. Quantitatively, this means unstable regions (e.g., Cluster 1, 50–70% $G_y$, ±60 $G_y$ spread) become more predictable in the global fit (less than ±25 $G_y$), while stable clusters (e.g., Cluster 6, RMSE=11.86) remain highly reproducible.

Overall, SoDip provides dual benefits: (a) cluster-level insight, preserving local variability and identifying irreproducible regimes, and (b) global stability, where borrowing information across clusters reduces fluctuations and tightens uncertainty bands. Quantitatively, unstable regions (e.g., Cluster 1, 150–200 $G_y$, ±60 spread) become more predictable in the CV fit (less than ±25 $G_y$), while stable clusters such as Cluster 6 remain highly reproducible. In sum, SoDip achieves accurate global calibration while directly addressing RIG's reproducibility challenge, enabling identification of reliable operating zones and flagging conditions where variability is unavoidable.

In conclusion, SoDip not only achieves accurate global calibration but also directly addresses the reproducibility challenge in RIG. By partitioning the dataset into clusters with distinct error profiles and variance structures, the framework allows scientists to distinguish stable operating zones from unstable conditions. This enables identification of fabrication recipes likely to yield reproducible outcomes while flagging those where variability is unavoidable, which is a key step toward overcoming the reproducibility problem in RIG.

**3.2. Analysis of RIG space using SoDip.**

RIG in this study is modelled as a 21-dimensional problem yet visualizing more than five dimensions in a single plot is impractical. Figure 6 therefore shows a 5D slice of the RIG operational-condition space: four independent variables: $D_{Dose}$, $D_{Temp}$, $D_{Time}$ and $D_{Monomer\_Conc}$, and one response, $G_y$. Plotting all 1,383 experimentally obtained $G_y$ values alongside a few SoDip-computed points $G_{y_*}$ ensures a smooth, continuous response surface.

The three orthogonal axes represent $D_{Dose}$ (x-axis), $D_{Monomer\_Conc}$ (y-axis) and $D_{Temp}$ (z-axis). $D_{Time}$ is encoded by a dotted mesh: bright yellow dots mark the maximum (24 h), while white dots mark the minimum (0.5 h), as indicated by the legend. Surface colour depicts $G_y$, and $G_{y_*}$ from dark blue (lowest) to dark red (highest).

Two "origin" points (A) and (B) on this surface represent the high-$G_y$ and the medium-$G_y$ regions respectively, each spawn subsidiary points under identical operating conditions but differing categorical descriptors:

- **Point A** ($D_{Dose}$ 45 kGy, $D_{Temp}$ 25 °C, $D_{Time}$ 22 h, $D_{Monomer\_Conc}$ 70 vol %)
  - $A_1$ (PS-g-ETFE; γ-ray pre-irradiation): $G_y$ = 408.01 %
  - $A_2$ (pVBC-g-PEEK; γ-ray simultaneous irradiation): $G_y$ = 83.01 %
  - $A_3$ (pVBC-g-FEP; β-particle pre-irradiation): $G_y$ = 81.01 %
  - $A_4$ (PS-g-PFA; γ-ray simultaneous irradiation): $G_y$ = 125.64 %
- **Point B** ($D_{Dose}$ 55 kGy, $D_{Temp}$ 40 °C, $D_{Time}$ 24 h, $D_{Monomer\_Conc}$ 80 vol %)
  - $B_1$ (pVBC/AN-g-ETFE; ion-beam simultaneous irradiation): $G_y$ = 236 %
  - $B_2$ (p1VIm-g-ETFE; γ-ray pre-irradiation): $G_y$ = 236 %

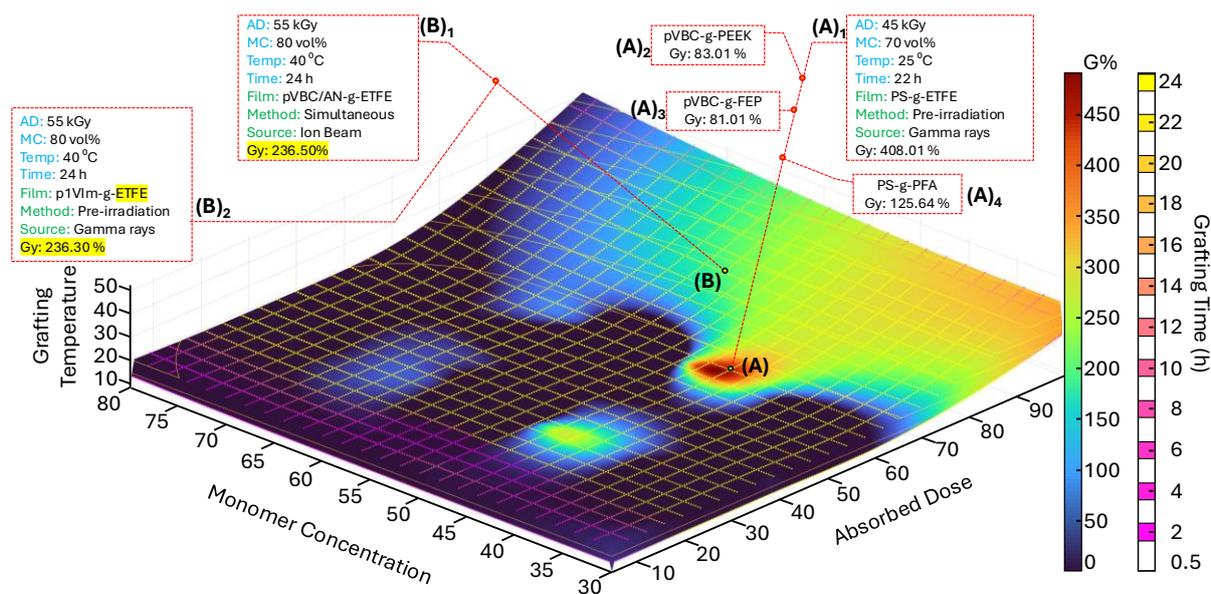

Figure 6. Response surface showing coupled effect of $D_{Dose}$, $D_{Temp}$, and $D_{Time}$ on $G_y$ within the RIG space as generated by SoDip.

These observations demonstrate that identical operational parameters can yield vastly different $G_y$ values when categorical descriptors (grafting method, radiation source, substrate/monomer combination) differ. The exceptionally high $G_y$ (408 %) of A$_1$ under moderate conditions contrasts sharply with A$_2$–A$_4$, illustrating that categorical factors can dominate performance. Conversely, B$_1$ and B$_2$ achieve the same $G_y$ despite differences in substrate and radiation type, highlighting complex, non-linear interactions between categorical and operational variables.

Relying solely on operational parameters can therefore be misleading. Instead, SoDip enables the generation of parameter subspaces grouped by categorical descriptors, allowing more precise, context-specific evaluation. By partitioning the design space into homogeneous subregions, we can mitigate confounding effects, avoid spurious correlations and improve predictive accuracy. The following sections develop this methodology, detailing how categorical grouping combined with multi-dimensional response surfaces can guide robust process optimization in radiation-induced grafting systems.

### 3.3. Analysis of RIG Subspaces using SoDip.

Based on our SoDip analysis of RIG space (Section 3.2), isolating subspaces by categorical descriptors is essential. In our dataset, ETFE-based films dominate (751 of 1,383 points), and within those, PS-g-ETFE appears most often (633 points; see S3 for the full film/monomer mapping). With a 70:30 split, 189 PS-g-ETFE points form the test set ample for assessing SoDip's predictions. We plotted the 633 PS-g-ETFE points (without augmentation) to generate a smooth response surface (Figure 7). The values for $G_y$ and $G_{y_*}$ were computed using SoDip and plotted as depicted in Figure 7 (in similar way as described in Figure 6). Owing to the copyrights restrictions by publishers we the results of the 189 test points couldn't be fully disclosed so, we just showed A-E were test validation results shown in green colour (shown as G%) taken from ref [1-7].

In Figure 7, the highest $G_y$ regions (reddish orange to dark red) cluster around $D_{Monomer\_Conc} \approx 40$ vol% across the full $D_{Dose}$ range (25–100 kGy), indicating strong sensitivity to monomer concentration at that level. Intermediate yields (bluish to yellowish green) span roughly 30–50 vol% $D_{Monomer\_Conc}$ across doses, persisting at $D_{Dose}$ = 60–100 kGy when $D_{Temp}$ is between 45 °C and 65 °C. We validated SoDip's predictability using six representative points (A–F) ref [1-7]:

**A** (25 kGy, 40 vol%, 50 °C, 40 h): $G_{y_*}$ = 215.01% vs. G% = 212.48% (Δ = 2.53)
**B** (40 kGy, 40 vol%, 45 °C, 60 h): $G_{y_*}$ = 219.97% vs. G% = 221.38% (Δ = 1.86)
**C** (60 kGy, 40 vol%, 50 °C, 60 h): $G_{y_*}$ = 222.73% vs. G% = 221.00% (Δ = 1.73)
**D** (70 kGy, 40 vol%, 55 °C, 60 h): $G_{y_*}$ = 280.51% vs. G% = 273.40% (Δ = 2.6)
**E** (100 kGy, 40 vol%, 65 °C, 60 h): $G_{y_*}$ = 430.99% vs. G% = 433.90% (Δ = 2.91)
**F** (100 kGy, 60 vol%, 60 °C, 60 h): $G_{y_*}$ = 320.44% vs. G% = 324.64% (Δ = 4.20)

Across these, the average deviation was 2.63%, and the maximum was 4.20%, demonstrating strong predictive performance.

This subspace analysis highlights monomer concentration as the dominant variable, with an optimal plateau at ~40 vol%. Beyond this, increasing monomer yields diminishing improvements, likely due to viscosity-limited diffusion or radical recombination [69–72]. Dose and temperature synergistically modulate $G_y$ within this $D_{monomer\_conc}$ window: increasing $D_{Dose}$ from 25 to 100 kGy steadily elevates $G_y$, but the marginal gain diminishes at temperatures above 60 °C, indicating thermal activation of side reactions or homopolymerization [69,73–75]. The slight overprediction at point D indicates higher uncertainty near the upper $D_{Dose}$-$D_{Temp}$ boundary, where radical lifetimes and monomer mobility compete. Success at the extreme point E confirms SoDip's robustness even under intense irradiation.

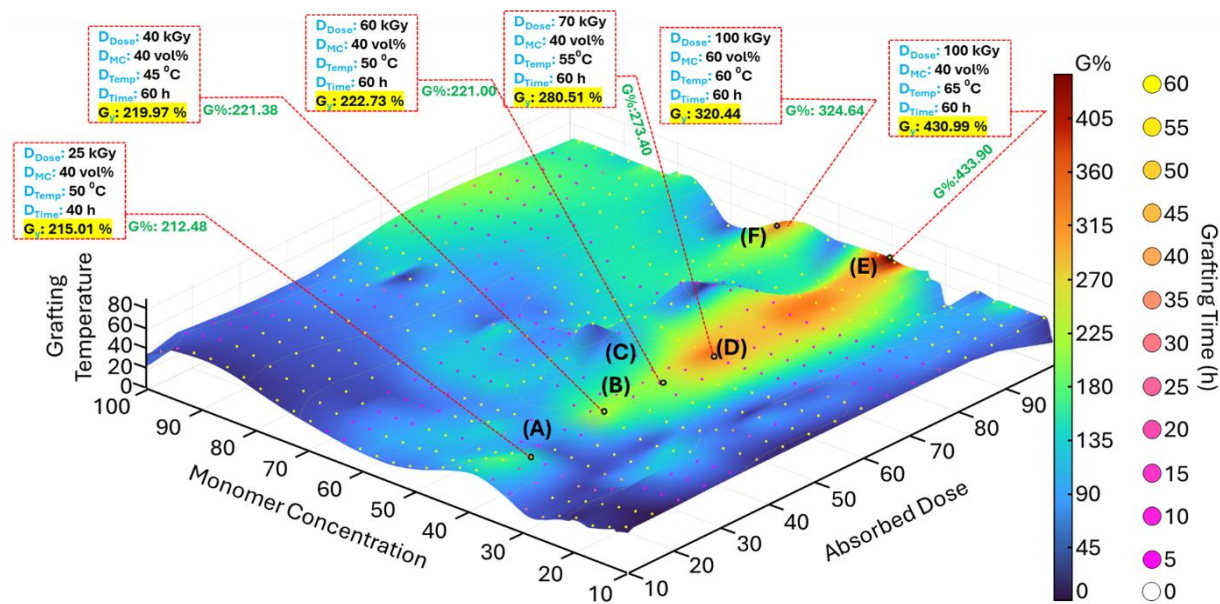

**Figure 7.** RIG subspace of PS-g-ETFE.

Having validated SoDips's predictability for PS-g-ETFE, more extreme cases are provided at supplementary information section where SoDip is examined using a less-populated subspace: PS-g-PTFE with too few points (62 points total; 18 in the test set) at Figure S3. Also, SoDip's ability to generate RIG subspaces for a chemically distinct RIG system such as poly(vinylbenzyl chloride)-g-poly(ether ether ketone) (pVBC-g-PEEK) is demonstrated at Figure S4.

### 3.4. Analysis of categorical descriptors

The effect of categorical descriptors such as $D_{Source}$, and $D_{solvents}$ on $G_y$ were also presented in this study. In this section we study the effect of solvents on $G_y$ for all grafted films presented in our dataset. Other categorical descriptors are found in the supplementary information section (Refer to Table S2 for data sample representing all descriptors).

Figure 8 presents the relationship between solvent identity and the $G_y$ for a series of grafted polymer films. In the central heat-map, individual grafted film types are arranged along the horizontal axis and solvents along the vertical axis. Marginal histograms accompany the heat-map: the top-histogram displays the total number of data points for each grafted film, and the right-hand histogram shows the total number of observations for each solvent.

The right-hand histogram showed that methanol is the most frequently employed solvent, with 320 data points. Also, Isopropanol, dichloromethane, benzene, and N,N-dimethylformamide each appear in 148–152 experiments. Remarkably, Water was used in 97 experiments, and 159 experiments were conducted without any solvent. On the top histogram, the grafted film occurrence showed that PS-g-ETFE is the most reported film in our dataset, appearing 633 times in the dataset. PS-g-PVDF and PS-g-FEP follow, with 187 and 157 observations, respectively. Moreover, PS-g-PFA and p1VIm-g-ETFE appear in 87 and 84 experiments, respectively. All other grafted films are reported between 1 and 33 times. This presentation highlights both the dominant solvent systems and the most extensively studied grafted films, facilitating comparison of solvent effects on grafting yield across the dataset.

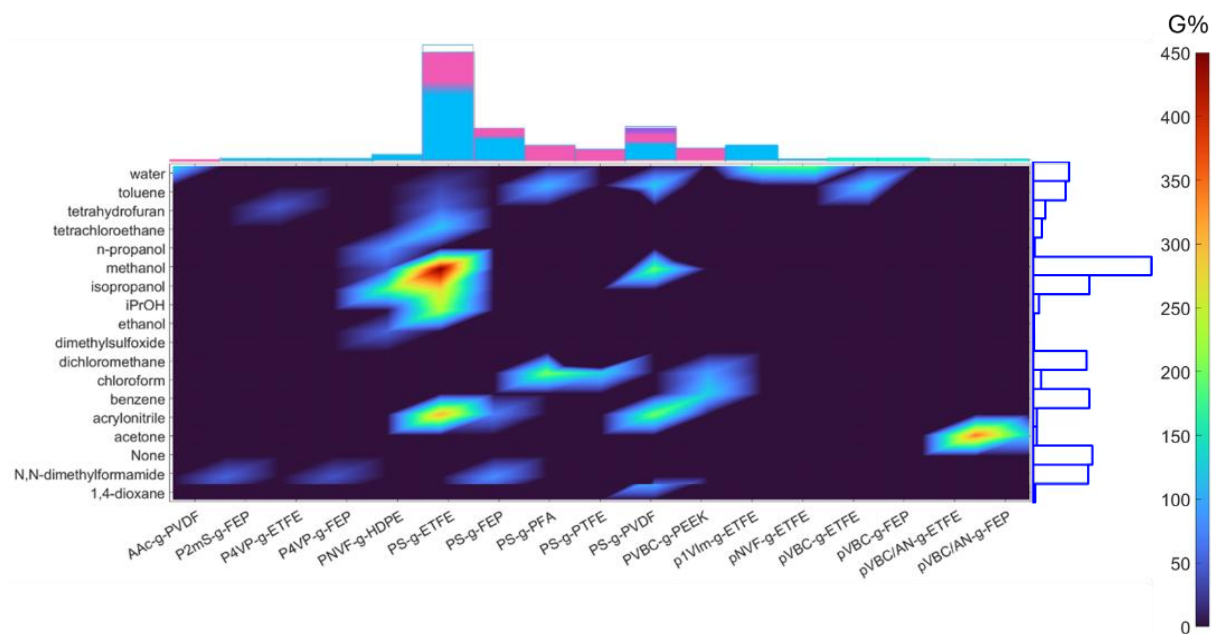

**Figure 8.** Effect of D$_{solvents}$ on the grafting yield for different grafted films.

Grafting yield data across various solvent–monomer combinations are summarized in Figure 8. In aqueous medium, acrylic acid achieved the highest grafting yield compared to N-vinylformamide. Where, $G_y$ for grafting acrylic acid onto PVDF (AAc-g-PVDF) reached 92 % yield, while $G_y$ for grafting 1-vinylimidazole onto ETFE (p1VIm-g-ETFE) and N-vinylformamide onto ETFE (pNVF-g-ETFE) was116.6 % and 120.3 %, respectively. In our dataset no other monomers were tested in water.

For non-polar aromatic solvent (toluene), styrene grafts (PS-g-ETFE, PS-g-PVDF) delivered moderate yields with $G_y$: 50.5 %, 82 % respectively, and PS-g-PFA reached $G_y$: 74.4 %. Also, grafting vinyl benzyl chloride grafts onto PEEK (pVBC-g-PEEK) achieved $G_y$: 28.9 % while grafting vinyl benzyl chloride-acrylonitrile comonomer onto FEP (pVBC/AN-g-FEP) exhibited $G_y$: 24.2 %, whereas $G_y$ for the pure pVBC-g-FEP reached 81.0 %. On other hand, Benzene swelling produced highest $G_y$ for grafting on ETFE, FEP and PEEK as for PS-g-ETFE ($G_y$: 212.5 %), PS-g-FEP ($G_y$: 62.0 %), and pVBC-g-PEEK ($G_y$: 122.1 %).

In chlorinated solvents, dichloromethane afforded excellent yields for PS-g-PFA ($G_y$:125.6 %) and PS-g-PTFE ($G_y$:85.9 %), whereas chloroform resulted in $G_y$: 83.2 % for pVBC-g-PEEK.

Protic alcohols (methanol, isopropanol) strongly promoted PNVF-g-HDPE ($G_y$: 52.5 % in methanol, and $G_y$: 112.2 % in i-PrOH) and PS-g-ETFE ($G_y$: 433.9 % in methanol, and $G_y$: 180.3 % in i-PrOH) which is relatively high compared to PS-g-ETFE in ethanol ($G_y$: 55.8 %).

Polar aprotic solvents DMF and acetone yielded modest PS-g-PVDF (83.1 % in DMF) and PS-g-PTFE ($G_y$: 35.98 % in acetone). In tetrahydrofuran (THF), P4VP-g-ETFE achieved $G_y$: 54.0 %, for PS-g-FEP $G_y$ was 56.5 %, and for PS-g-PTFE grafting was low ($G_y$: 10.9 %). Single-point trials of other solvents such as n-propanol, dimethyl sulfoxide, and 1,4-dioxane gave yields of 65.8 %, 5.3 %, and 43.6 %, respectively, for those few grafts attempted.

The data reveal strong solvent effects on grafting efficiency, tied closely to solvent polarity and monomer–solvent compatibility. Water, as a highly polar medium, excels for ionizable monomers (acrylic acid, N-vinylformamide, vinylimidazole), likely due to enhanced monomer solubility and favourable radical stabilization, which promotes graft chain growth [76–79]. In contrast, nonpolar aromatic solvents (toluene, benzene) maximized yields for hydrophobic styrene grafts (PS-g-ETFE, PS-g-PFA, PS-g-PVDF) by promoting base film swelling and facilitating monomer diffusion into the polymer matrix. Chlorinated solvents (dichloromethane, chloroform) combine moderate polarity with good substrate swelling [ref], delivering the highest yields overall for PS-g-PFA (125.6 %) and PS-g-PTFE (85.9 %).

Protic alcohols (methanol, i-PrOH) showed exceptional performance for N-vinylformamide grafting onto HDPE (PNVF-g-HDPE), indicating hydrogen-bonding interactions accelerate graft initiation and propagation for amide-type monomers, but were less effective for nonpolar styrene. Polar aprotic media (DMF, acetone) gave moderate yields for PS grafts, suggesting sufficient monomer solubility but limited substrate swelling. Single-point trials in THF, n-propanol, DMSO, and 1,4-dioxane confirm that while some grafting is possible, comprehensive screening is needed to identify optimal conditions. Overall, tailoring solvent selection to the dual requirements of monomer polarity and base-film swelling is key to maximizing grafting efficiency across diverse polymer–monomer systems.

## 4. Conclusion

In this work, we have introduced SoDip, a hybrid, Bayesian non-parametric framework that synergistically combines Transformer-based text encoding, multimodal tabular regression, GPR, DPMM clustering, and Bayesian optimization to predict RIG grafting yields with high accuracy and calibrated uncertainties. In our evaluation, such architecture enabled SoDip to overcome key limitations of previous methods that often ignored unstructured textual metadata, lacked mechanisms to partition heterogeneous data regimes, and did not provide calibrated confidence intervals. By leveraging DPMM to partition complex, heterogeneous datasets into locally homogeneous regimes and applying GPR within each cluster, SoDip effectively models nonlinear, cluster-dependent relationships and quantifies predictive confidence even when the true number of latent regimes classes is unknown. Besides, it attains flexibility and robustness in practice, making it particularly well suited for RIG, RAFT, and catalytic systems where reproducibility is imperfect, and uncertainty quantification is critical. Our stepwise evaluation demonstrates that:

1. Meta-variable learning

The agnostic-learning block (DeepSeek-R1 + TabNet + XGBoost) captures 89-92% of variance in the intermediate meta variable $\mathcal{Z}_n$ (MAE: 2.73-3.09%), outperforming each individual component.

2. Nonlinear regression with uncertainty

Standalone GPR on the stacked-generalization output achieves $R^2$: 0.72 (MAE: 3.0), confirming its ability to model complex nonlinearities and to expose increased uncertainty at extreme responses.

3. Cluster-aware modelling

DPMM-clustered GPR dramatically enhances both fit and generalization R²: 0.96 (MAE: 4.5 % of $G_y$) in cross-validation and R²: 0.93 (MAE: 5.8 %) on held-out tests by capturing local heterogeneities and heteroscedastic noise. Besides, DPMM reduced unexplained variance from 28% to 4% (≈85% reduction), indicating the model now captures much more of the underlying data structure and variability.

4. Design-space insights

Five-dimensional RIG-space visualizations and subspace analyses (e.g., PS-g-ETFE, PS-g-PTFE, pVBC-g-PEEK) showed that categorical descriptors can drive yield variations of several hundred percent under identical conditions, highlighting the need for clustering by these factors.

5. Extrapolative validation

Predictive subspace generation via LHC sampling confirms SoDip's extrapolative power (average deviations ≤ 5 units) and reveals unreported high-yield regions.

Together, these findings show that (i) explicitly incorporating base-film origin and other categorical descriptors is essential to resolve reproducibility challenges in RIG, and (ii) SoDip delivers state-of-the-art predictive accuracy with calibrated uncertainty estimates that flag low-data-density or high-variability regions. By enabling rapid, data-driven optimization of dose, temperature, reaction time, and monomer concentration across diverse polymer–monomer systems, SoDip advances reproducible, efficient development of functional grafted membranes and electrolytes. Future work will extend this SoDip to other types of RIG like emulsion RIG, our framework could be extended to predict grafting rate in relation to micelles size, and micelles growth and decay rates. Also, study should be extended to include different base polymer types like resins and nanofibers.

**Supplementary Information: Hierarchical Stacking Optimization Using Dirichlet's Process (SoDip): Towards Accelerated Design for Graft Polymerization**

Each stage of SoDip is designed to manage memory efficiently (using batch processing, CUDA cache clearance, and mixed precision) and optimize the overall model performance. This modular approach makes it easier to diagnose performance issues and adjust individual components as needed. It maintains an optimized model pipeline with batch processing and memory handling that perform transformer-based modelling over GPU, while the probabilistic modelling runs mostly on CPU. Below is a detailed, breakdown and explanation of the workflow of SoDip. The purpose of each section, the role of key functions, and how data flows through the model pipeline will be also described.

**Data preparation**

Began with three different input types: textual embeddings, numeric discrete data, and numeric continuous. Textual data were stored as an array of strings, composition data as a numeric array of float64 values with 27 features per datapoint, and regression labels as float64 column vector. Each row of composition data was converted into a space-separated string, which was concatenated with the corresponding textual label. This combined text served as input to the DeepSeek transformer model. Regression labels were reshaped to match the input format for TabNet ($n_{samples}$,1). The dataset was then divided into training, validation, and testing sets by first isolating the test set, followed by splitting the remaining data into training and validation subsets.

**Transformer model and tokenizer setup**

DeepSeek-R1-Strategy-Qwen-2.5-1.5b-Unstructured-To-Structured model was loaded locally in half-precision (torch.float16) to reduce memory usage and improve inference speed. The model was configured to output hidden states, and the tokenizer was loaded with the end-of-sequence token set as padding. A custom batch processor was implemented to iterate through text and regression data in batches. Each batch of text was tokenized with padding and truncation (maximum sequence length of 64) before being passed through the transformer model on the GPU with mixed-precision inference. The hidden states from the last layer were retrieved and stored.

**Attention mechanism**

Was routinely applied by computing a scaled dot-product between hidden states to obtain attention weights. These were normalized with a SoftMax function and used to calculate a weighted sum, producing an attention-based representation. The regression data together with the attention-based features were stored in a tensor. This tensor will be detached, and moved to the CPU, converted to float32, and reshaped into a 2D array for each batch. Memory cleanup was performed using "torch.cuda.empty_cache()" and Python's garbage collector. The final output of this stage is a NumPy array containing processed features, ready for scaling and subsequent modelling.

**Feature scaling and dimensionality reduction**

Feature scaling is applied to the processed training, validation, and test inputs using scikit-learn's StandardScaler. This ensures that features contributed equally during training. To reduce dimensionality and computational cost, Principal Component Analysis (PCA) was applied. The number of principal components was limited to the minimum of 500, the number of features, and the number of training samples to mitigate overfitting.

**Training the TabNet regressor**

The reduced features were then used for training the TabNet regressor. The TabNet model was initialized with predefined hyperparameters, including the Adam optimizer, learning rate, and learning rate scheduler. Data were forced to float32 format when necessary. The model was trained on the PCA-transformed training features and regression targets, with validation data used for monitoring performance. Performance metrics included RMSE and MAE, and early stopping was enabled using a patience parameter.

**Feature extraction from TabNet**

Following training, feature extraction from TabNet was performed using the get_tabnet_features function, which applied model.predict() on PCA-reduced data. These extracted predictions served as high-level representations of the inputs. A new instance of StandardScaler was then used to rescale TabNet features across training, validation, and test sets to ensure standardized distributions prior to the next modeling stage.

**Hyperparameter optimization for XGBoost**

Next, hyperparameter optimization for XGBoost was carried out. A parameter space was defined that included the number of trees, learning rate, maximum tree depth, sampling ratios (subsample and colsample_bytree), and complexity parameters (min_child_weight and gamma). An objective function mapped candidate hyperparameters to model training and evaluation on the validation set, with RMSE as the metric. The gp_minimize function from scikit-optimize was used to explore the hyperparameter space across 50 iterations, and the best configuration was retained.

The final XGBoost model was then instantiated using the optimized hyperparameters and trained on the scaled TabNet features from the training set. Predictions were generated for the test set, and RMSE values were computed for training, validation, and testing splits to provide a comprehensive evaluation. For reproducibility, the trained TabNet model was saved as a compressed file (tabnet_model.zip) in the working directory, which was logged for reference.

**Visualization and evaluation**

Finally, visualization and evaluation metrics were performed. Scatter plots comparing predicted versus actual values were generated for validation and test sets to allow visual inspection of predictive accuracy. Alongside RMSE, additional metrics such as MAE and the coefficient of determination ($R^2$) were computed and reported to provide a more complete assessment of model performance.

XGBoost, being a tree-based model, does not inherently require inputs to be normalized or to be input in form of probabilities. However, the softmax-processed attention output helps in two keyways:

a) Improved Feature Quality: It refines the raw transformer embeddings into a representation that highlights important contextual information. This "denoised" and enriched representation, when further processed (e.g., via PCA and TabNet), creates a feature set that can improve the predictive performance of XGBoost.

b) Stability and Consistency: The normalization inherent in softmax can contribute to more stable training of intermediate models (TabNet) by ensuring that the feature magnitudes are within a controlled range. This, in turn, can help downstream models like XGBoost perform better on a more consistent feature space.

Thus, in brief, softmax plays a crucial role in the attention mechanism to create focused and normalized representations from the transformer's output. Although XGBoost doesn't inherently require softmaxed inputs, the improved quality of the features generated by this attention mechanism contributes positively to the overall performance of the final regression model as proved experimentally while designing and constructing the SoDip. A comparison of the performance with and without XGBoost is presented in figures S1.

Figure S1(A) presents the performance of the SoDip pipeline without the inclusion of XGBoost, where the model achieved a validation RMSE of 31.182 and a test RMSE of 34.073. Corresponding MAE values were 15.611 (validation) and 17.943 (test), while the coefficient of determination ($R^2$) reached 0.572 and 0.568 for validation and test sets, respectively. These results indicate moderate predictive accuracy but also highlight the presence of notable residual errors. In contrast, Figure S1(B) shows the results obtained after integrating XGBoost with the TabNet-derived features, leading to a substantial improvement in predictive performance. The validation RMSE decreased to 10.719 and the test RMSE to 11.177, with MAE values dropping to 5.494 and 5.917 for validation and test sets, respectively. Likewise, $R^2$ values increased to 0.752 (validation) and 0.728 (test), reflecting stronger explanatory power and a more robust generalization across datasets. Collectively, these comparisons demonstrate that coupling TabNet with XGBoost considerably enhanced both accuracy and reliability, underscoring the benefit of the hybrid modeling approach adopted in SoDip.

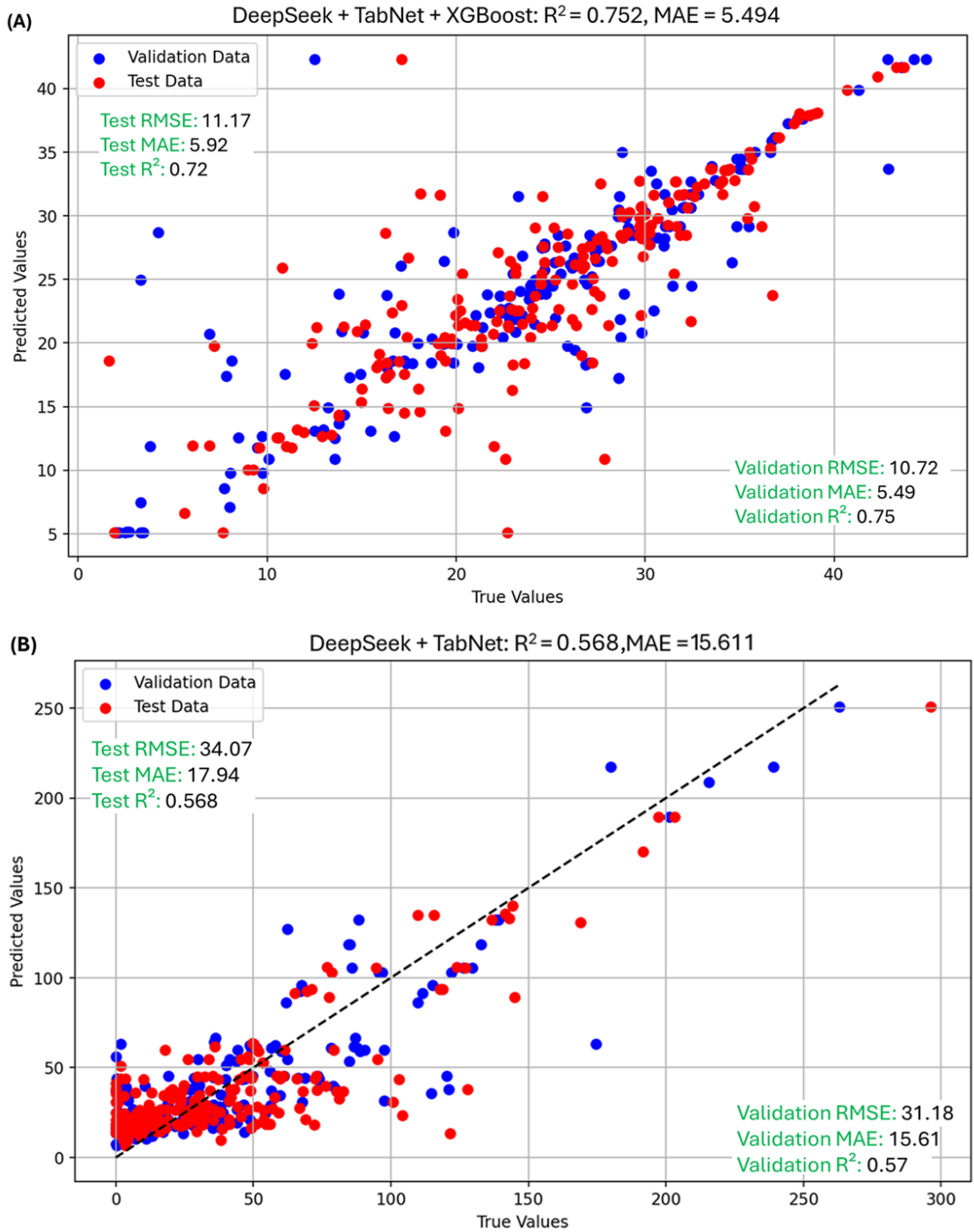

Figure S1(A) presents a parity plot showing true vs of SoDip's predictions without the inclusion of XGBoost, Figure S1(B) shows the parity plot for SoDip's predictions obtained after integrating XGBoost with the TabNet-derived features.

**Cluster-level breakdown**

The DPMM achieved strong global calibration, with 95% of observed values lying within the 95% prediction interval (PI). Importantly, the cluster-level breakdown highlights distinct fabrication regimes with widely varying reproducibility characteristics Table S1 highlights remaining representative clusters that not discussed in the main text.

At the low-yield end (<170 $G_y$), Cluster 8 stands out as the most reliable regime ($R^2$=0.98; RMSE=8.44), with a flat, bit noisy and symmetric distribution indicating exceptional reproducibility as all $G_y$ values (at this range) are equally represented. By contrast, Cluster 3 ($R^2$=0.33; RMSE=17.30) exhibits weak predictive power and noisy variability, making low-yield outcomes in this range irreproducible despite similar $G_y$ values, which is attributed to its sharp, symmetric distribution where very specific values of $G_y$ dominate the cluster.

In the mid-yield range (200–300 $G_y$), Cluster 7 exhibited moderate fit ($R^2$=0.66; RMSE=27.97), highlighting that not all recipes in this range are equally represented. Despite the proper distribution of $G_y$ values within the cluster, the small size of cluster contributed to the reduced fit quality.

Finally, Clusters 2, 4, and 5 capture mixed regimes. While Cluster 2 ($R^2$=0.88; RMSE=20.62) and Cluster 4 ($R^2$=0.83; RMSE=20.87) provide nearly high predictive reliability, despite of the error levels. The large clusters size compensates the biased distribution of $G_y$ values. while Cluster 5 ($R^2$=0.12; RMSE=39.12) performs very poorly due to high skewness and small cluster size, indicating conditions where reproducibility is essentially unattainable.

In summary, the DPMM partitions the dataset into reproducibility classes: highly stable (Clusters 6, 8), nearly stable (Clusters 1, 2, 4), and unstable (Clusters 3, 5, 9). This regime-aware structure allows fabrication scientists to identify operating zones where outcomes can be replicated with high confidence, while clearly flagging conditions where reproducibility is inherently compromised.

**Table S1:** Cluster-Level Model Performance and Reproducibility Characteristics

| Cluster | n | $R^2$ | RMSE | $G_y$ Range | Distribution Shape | Reproducibility Assessment |
|---|---|---|---|---|---|---|
| 2 | 119 | 0.88 | 20.62 | 0–280 | 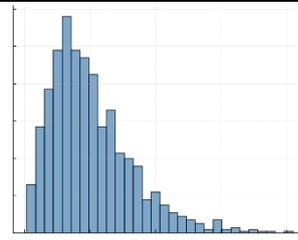 Broad, skewed | Nearly stable: acceptable fit, but error level indicates reproducibility challenges. |
| 3 | 168 | 0.33 | 17.30 | <120 | 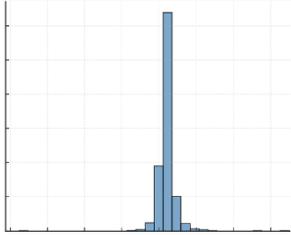<br>Narrow, symmetric | Unstable: weak fit (low $R^2$), poor reproducibility at low yields. |
| 4 | 173 | 0.83 | 20.87 | <350 | 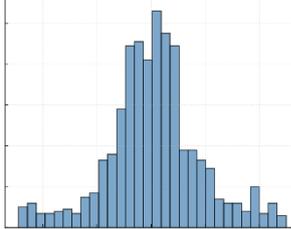<br>Noisy, irregular | Nearly stable: some predictive power but wide variability reduces reproducibility. |
| 5 | 53 | 0.12 | 39.12 | <300 | 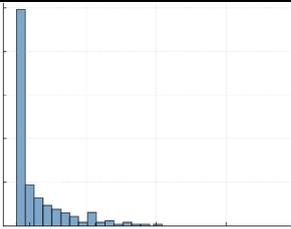<br>Skewed, heavy-tailed | Unstable: very poor fit ($R^2$=0.12), outcomes unpredictable. |
| 7 | 60 | 0.66 | 27.97 | 200–300 | 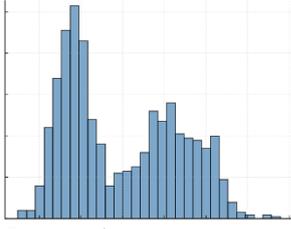<br>Broad, irregular | Moderately stable: small cluster size, moderate fit and high error undermine reproducibility. |
| 8 | 188 | 0.98 | 8.44 | <170 | 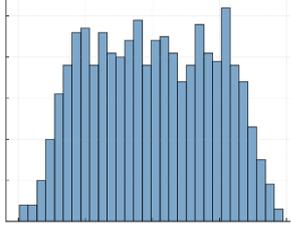<br>Flat, noisy | Exceptionally stable: best-performing cluster, highly reproducible low-yield regime. |

**Table S2:** Data set sample representing (a) Categorical descriptors, (b) Numerical descriptors

a. Categorical descriptors

| Data index | Film name | Monomer | Base polymer | Base polymer structure | Base Polymer supplier | Additive | Grafting type | Irradiation type | Solvent type |
|---|---|---|---|---|---|---|---|---|---|
| 14 | PS-g-ETFE | Styrene | ETFE | Film | Goodfellow | None | pre-irradiation | Electron-Beam | methanol |
| 183 | PS-g-PTFE | Styrene | PTFE | Film | Hanmi Rubber and Plastics Co | none | simultaneous irradiation | gamma | N,N-dimethyl formamide |
| 283 | PS-g-ETFE | Styrene | ETFE | Film | Goodfellow | None | pre-irradiation | Electron-Beam | methanol |
| 712 | PS-g-FEP | Styrene | FEP | Film | DuPont | None | simultaneous irradiation | Electron-Beam | None |
| 816 | PS-g-PVDF | Styrene | PVDF | Film | Plastpolimer | None | simultaneous irradiation | helium ion beam | methanol |
| 861 | PS-g-PVDF | Styrene | PVDF | Film | Goodfellow | sulfuric acid | pre-irradiation | Electron-Beam | N,N-dimethyl formamide |
| 1195 | PVBC-g-PEEK | vinylbenzyl chloride | PEEK | Film | Goodfellow | None | simultaneous irradiation | gamma | 1,4-dioxane |
| 1293 | pVBC/AN-g-ETFE | vinylbenzyl chloride | ETFE | Film | Nowofol GmbH | None | pre-irradiation | Beta | acrylonitrile |
| 1382 | p1VIm-g-ETFE | 1-vinyl imidazole | ETFE | Film | Nowofol GmbH | ferrous sulfate | pre-irradiation | Electron-Beam | water |

b. Numerical descriptors

| Data index | Base Polymer size | Solv. conc. | Add. conc. | Absorb. Dose | Graft Temp. | Graft Time | Monomer conc. | Degree of grafting | MW Film (repeating unit) | MW monomer | MW Solve. | MW Add. |
|---|---|---|---|---|---|---|---|---|---|---|---|---|
| 14 | 125 | 0.4 | 0 | 100 | 60 | 36 | 99.6 | 77.0 | 100.08 | 104.15 | 32.04 | 0 |
| 183 | 80 | 50 | 0 | 30 | 25 | 24 | 50 | 3.09 | 100.02 | 104.15 | 73.09 | 0 |
| 283 | 125 | 60.2 | 0 | 100 | 60 | 48 | 39.8 | 315 | 100.08 | 104.15 | 32.04 | 0 |
| 712 | 125 | 0 | 0 | 50 | 30 | 2 | 100 | 4.38 | 250.04 | 104.15 | 0 | 0 |
| 816 | 50 | 50 | 0 | 1500 | 60 | 6 | 50 | 18 | 64.02 | 104.15 | 32.04 | 0 |
| 861 | 50 | 77.6 | 10 | 100 | 60 | 36 | 12.4 | 27.64 | 64.02 | 104.15 | 73.09 | 98.08 |
| 1195 | 50 | 30 | 0 | 40 | 25 | 20 | 70 | 43.61 | 288.3 | 152.62 | 88.11 | 0 |
| 1293 | 50 | 30 | 0 | 50 | 60 | 12 | 70 | 130.36 | 100.08 | 152.62 | 53.06 | 0 |
| 1382 | 50 | 11.11 | 1.39 | 100 | 60 | 72 | 87.5 | 80.09 | 100.08 | 94.11 | 18.02 | 151.91 |

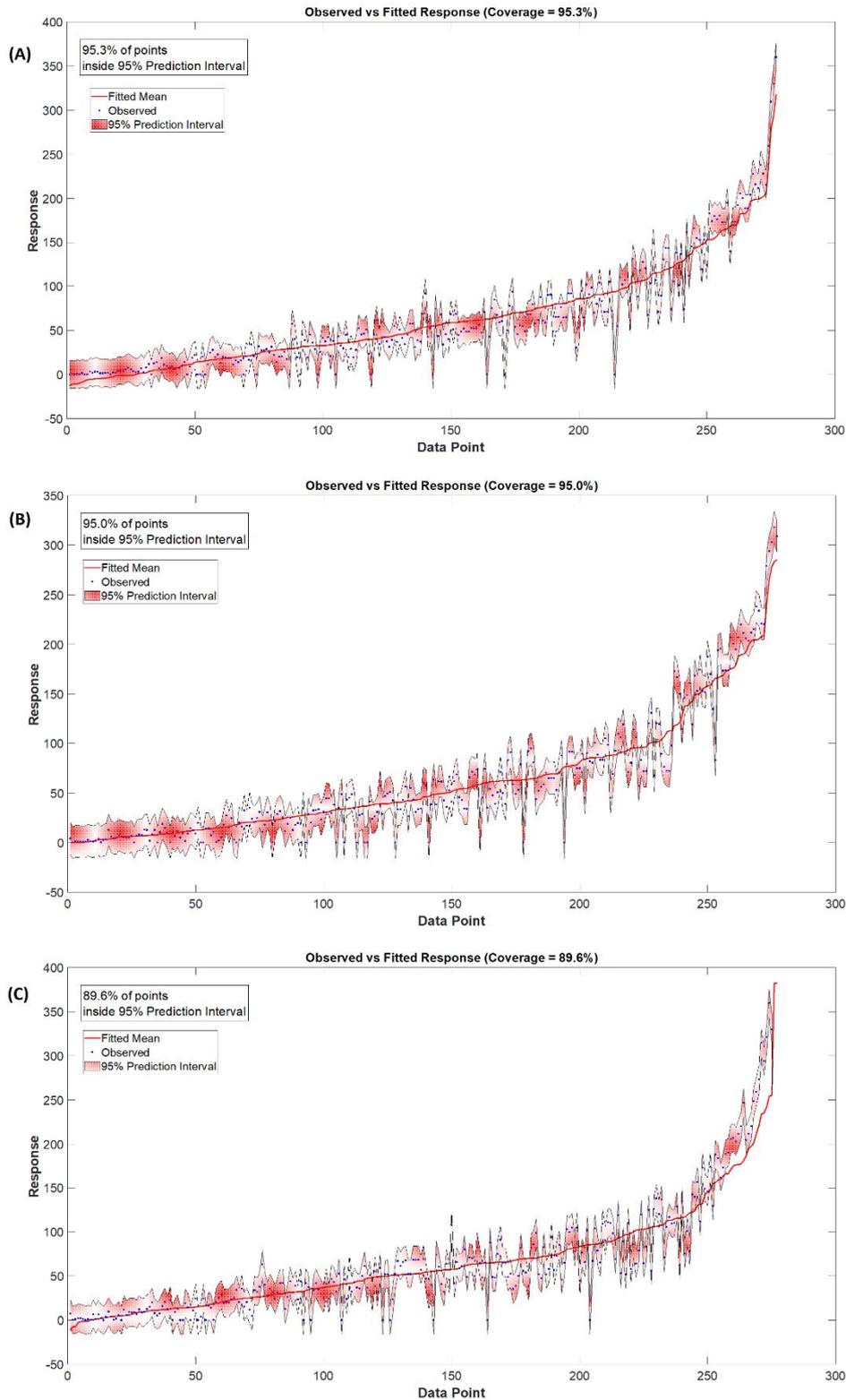

**Figure S2 (A–C):** represents the fitted response for the entire RIG dataset under six representative hyperparameter configurations, with posterior predictive means (red curves), observed responses (black markers), and 95 % prediction intervals (red shaded bands) obtained from GPR within DPMM clusters.

Figure S2 (A–C) presents the fitted response for the entire RIG dataset under six representative hyperparameter configurations, with posterior predictive means (red curves), observed responses (black markers), and 95 % prediction intervals (red shaded bands) obtained from GPR within DPMM clusters. The key hyperparameters considered are: α (Dirichlet concentration), K (maximum cluster count), Scale (covariance scaling), and $v_0$ (degrees of freedom for the prior covariance). Across all settings, the predictive mean generally tracks the empirical observations closely over the central response range (0–200 %), while the prediction intervals widen at both the low (< 20 %) and high (> 250 %) extremes, reflecting data sparsity and the model's ability to adapt to heteroscedastic regimes. Notably, the steep upward curvature beyond 250 % is consistently captured by both the fitted mean and the widening intervals.

Closer inspection highlights distinct trade-offs across the three panels. Figure S2 (A) is result of hyperparameter configuration: α = 0.1, K = 8, Scale = 1, $v_0$ = d+1 yields relatively balanced fitting with reasonable central fit and tail coverage a considerable underfitting is observed at $G_y$ values < 20%, while predicted mean failed to match observations of $G_y$ values >300%. Figure S2 (B), represents hyperparameter configuration: α = 0.01, K = 12, Scale = 3, $v_0$ = d+3. Such configuration improved the predictability at $G_y$ values < 20%. It also, provides broader prediction intervals with stronger coverage at $G_y$ values >300%, yet the predicted mean still failed to meet observations. Thus, configurations in figures S2 (A) and (B) represent balanced trade-off, with adequate coverage in the tails and moderately tight central bands. Figure S2 (C) depicts the outcome predictions of hyperparameter configuration: α = 0.001, K = 12, Scale = 5, $v_0$ = d+5. For this configuration, the predicted mean managed to trace mostly the entire range of observations, especially at high-variance regions, particularly in the upper tail, however, this configuration failed to represent $G_y$ values 200-325%. In figures S2 (A) and (B) 95% of observations was traced by predicted means as denoted by coverage % , however coverage dropped to 89% for configuration presented at figure S2 (C), so, this compromised the coverage for the sake of improving the predictability of extreme $G_y$ values.

These behaviours highlight the value of combining DPMM clustering with local GPR. Unlike a single global GPR, which risks over or under-confidence, SoDip allows kernel hyperparameters to vary across clusters. This enables clusters in data-rich regions to maintain sharp uncertainty bands, while clusters in sparse regions express larger epistemic uncertainty. The resulting intervals capture both epistemic uncertainty (growing with cluster sparsity) and aleatoric uncertainty (intrinsic fabrication noise in RIG). By explicitly modelling these two sources of uncertainty, the framework provides a principled way to identify parameter regimes where fabrication outcomes are inherently variable, thereby guiding more reliable process optimization.

**SoDip's performance at a less-populated subspace: PS-g-PTFE with too few points**

Having validated SoDips's predictability for PS-g-ETFE, more extreme cases are provided here where SoDip is examined using a less-populated subspace: PS-g-PTFE with too few points (62 points total; 18 in the test set). Figure S3 contains group of points A-E, where materials produced by recipes points A-C has matching combination in our test set (189 test points), points D and E showed predicted points that have good $G_y$ although we didn't find similar covered in literature. Due to copyrights and permission provided by publishers we can't show the results of the 189 test points so, we just showed A-C were test validation results shown in green color (shown as Gexp). It is evident that, the high to medium values of $G_y$ are distributed over wide area of $D_{Dose}$ and $D_{monomer\_conc}$ surface rather than being localized at the maxima of both descriptors. Namely, high occurred at $D_{Dose}$ in range of 16 - 40 kGy, and $D_{monomer\_conc}$ in range of 16 - 80 vol.%. When we focus on the $G_y$ range above 79 % (up to 94%), shown by conspicuous red colours from blight to dark red to show the $G_y$ ranges from 79 to 84 %. Especially, the dark red (high $G_y$s) are mainly located at peak areas, while cyan to light blue ($G_y$ in range of 24 - 14 %) dominates the centre of the plot, while the dark blue to black ($G_y$ in range of 4 - 0 %) exist at around edges of the plot. This is a visual indication that greater than 79 % occur at about room temperature ranges at the $D_{Dose}$ - $D_{monomer\_conc}$ surface. Exact values for points A-E are:

Figure S3 illustrates the distribution of $G_y$ across a design space defined by $D_{Dose}$, $D_{monomer\_conc}$, and $D_{Temp}$. Points A–C, corresponding to experimentally validated synthesis conditions $G_y$, exhibit close alignment between predicted and observed values (Δs = 3.57, 6.54, and 5.07, respectively). Notably, point B ($D_{Dose}$: 25 kGy, $D_{monomer\_conc}$: 60 vol%, $D_{Temp}$: 25°C) achieved a high $G_y$ of 79.33% ($G_{y_*}$: 85.87%), highlighting optimal performance within intermediate parameter ranges. Points D ($G_{y_*}$: 60.31%) and E ($G_{y_*}$: 79.71%) represent novel predictions not previously reported in the literature, with grafting conditions ($D_{Dose}$: 16–40 kGy, $D_{monomer\_conc}$: 16–38 vol%, $D_{Temp}$: 19–20°C) falling within broader operational windows.

The $G_y$ landscape reveals that high to medium yields ($G_y \geq 79\%$) are distributed across a wide $D_{Dose}$ (16–40 kGy) and $D_{monomer\_conc}$ (16–80 vol%) range rather than being confined to the maxima of either descriptor. High $G_y$ values (79–94%, dark red regions) predominantly occupy peak areas of the parameter space, while moderate yields (24–14%, cyan to light blue) dominate the central regions. Minimal yields (4–0%, dark blue to black) are localized at the periphery, particularly at extreme $D_{Dose}$ and $D_{monomer\_conc}$ values. Significantly, the highest $G_y$ values correlate with near-ambient temperatures (19–25°C), as exemplified by points A–E ($D_{Temp}$: 19–25°C).

The observed inverse correlation between maximal $G_y$ and the extreme values of $D_{Dose}$/$D_{monomer\_conc}$ suggests that optimal gelation does not require excessively high radiation doses or monomer concentrations. Instead, peak yields occur within intermediate ranges ($D_{Dose}$: 16–40 kGy; $D_{monomer\_conc}$: 16–80 vol%), implying a balance between monomer activation and crosslinking efficiency. This broad operational window enhances practical applicability, offering flexibility in parameter selection without compromising performance. The spatial distribution of $G_y$ further underscores the synergistic role of ambient temperature (19–25°C) in stabilizing reaction kinetics, as evidenced by the high yields of points B, D, and E under near-room-temperature conditions. It also, demonstrates that optimal grafting does not require maximized dose or monomer concentration; rather, high yields emerge over an extended "process window" of moderate $D_{Dose}$ and $D_{monomer\_conc}$ values. The model's ability to map these nonlinear interactions offers practical guidance for experimental design, enabling researchers to target moderate $D_{Dose}$/$D_{monomer\_conc}$ combinations that achieve ≥79 % grafting efficiency without resorting to extreme process settings.

The predictive capability of the model is validated by the strong agreement between calculated $G_y$ and experimental $G_y$ for points A–C (deviation ≤ 6.54%). while the novel predictions at points D and E, though lacking experimental validation, showcase its extrapolative capability to uncover high-performance not yet reported grafting regimes (e.g., $D_{Dose}$: 16 kGy, $D_{monomer\_conc}$: 38 vol%) that merit further investigation. Their predicted high $G_y$ values (60–80%) align with trends identified in the parameter space, suggesting robustness in the SoDip's extrapolation. However, the absence of literature precedents for these points highlights a knowledge gap in low-$D_{monomer\_conc}$, moderate-$D_{Dose}$ regimes, warranting targeted experimental studies, and emphasizes the capability of SoDip for accelerating discovery of new optimal conditions. Future work should integrate temperature and time directly into the surface model to refine predictions further and to facilitate multiobjective optimization of grafting conditions.

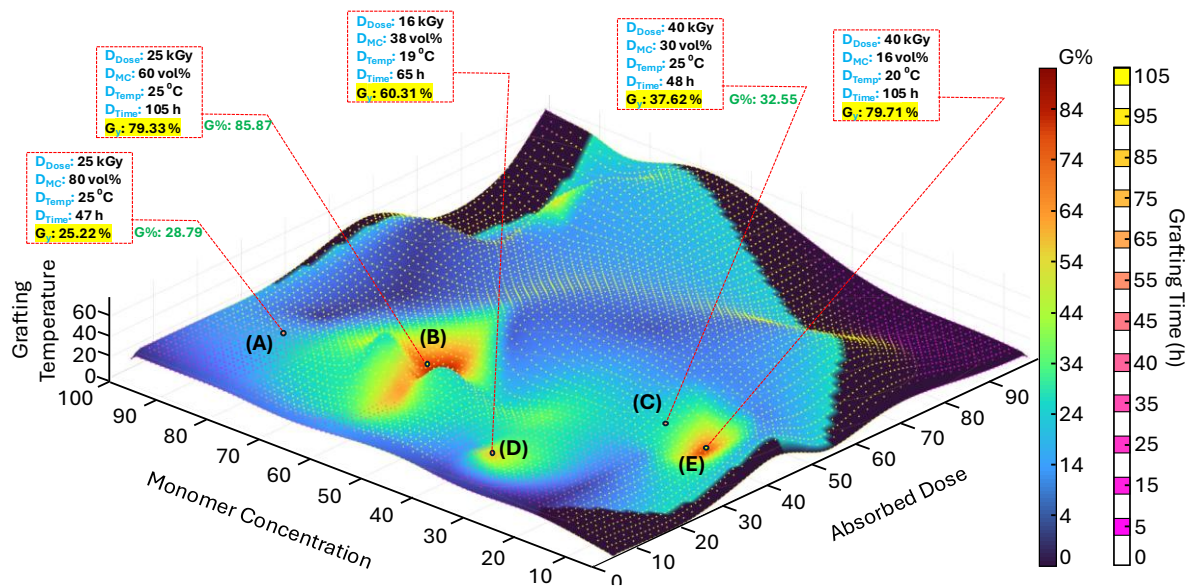

**Figure S3.** RIG subspace of PS-g-PTFE.

Based on the analysis of RIG space using SoDip (section 3.2), we found that the combination of $D_{Dose}$ and $D_{monomer\_conc}$ drastically affect the highest $G_y$s, i.e., $G_y$ ($D_{Dose}$, $D_{monomer\_conc}$) $_{max}$, whereas the $D_{Temp}$ and $D_{Time}$ shows relatively simple monotonical increases of $G_y$.

Now we demonstrate SoDip's ability to generate RIG subspaces for a chemically distinct RIG system poly(vinylbenzyl chloride)-g-poly(ether ether ketone) (pVBC-g-PEEK). Unlike the previously studied PS-g-ETFE, PEEK is a non-fluorinated, highly aromatic, thermally robust backbone, and VBC introduces reactive benzyl–chloride functionalities rather than inert styrenic vinyl groups. Moreover, only 68 experimental data points for pVBC-g-PEEK are available in our dataset, imposing a stringent test of the model's generalizability and robustness when extrapolating beyond its original training domain. Since that, 68 datapoints are not sufficient to generate smooth surface plot. We used LHC to generate 2000 RIG scenarios covering the observed minima and maxima of the 68 pVBC-g-PEEK points. The predicted $G_y$ values were obtained using SoDip and subsequently plotted alongside the predictions for the 68 test points, as illustrated in Figure S4.

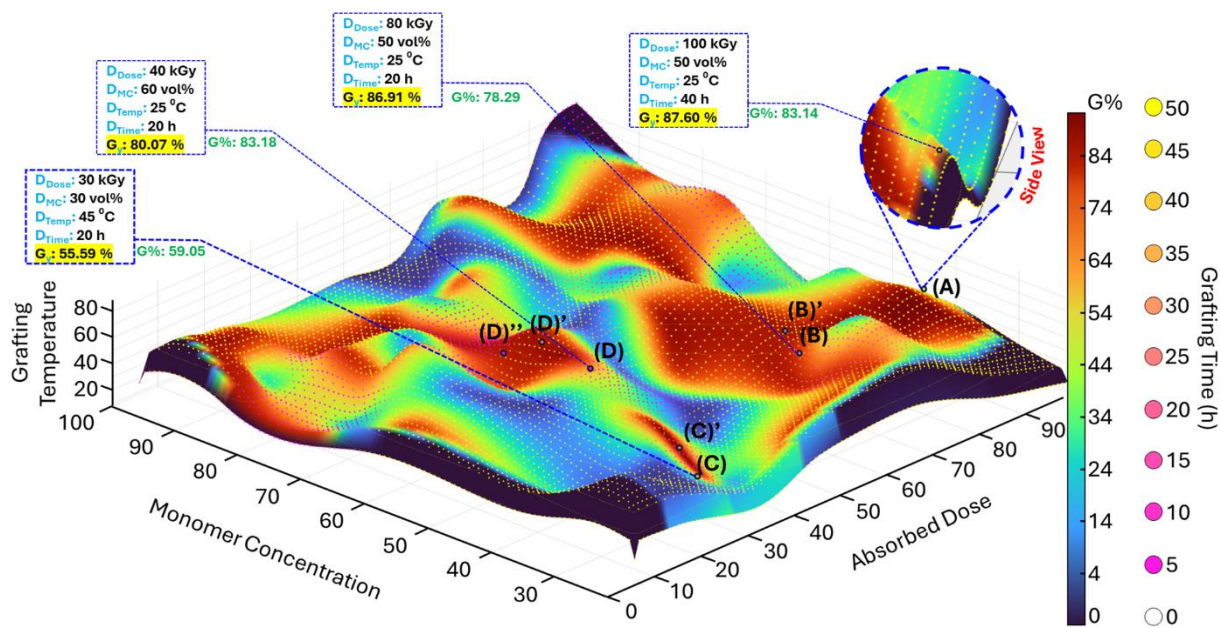

**Figure S4.** RIG subspace of pVBC-g-PEEK.

Figure S4 depicts the SoDip-predicted $G_{y_*}$ and the corresponding experimental observations $G_y$ on a three-dimensional response surface for the pVBC-g-PEEK subspace of RIG. Segregated islands where $G_y > 84\%$ (Cornell red to dark red) pervade the topology, indicating that the model predicts very high yields (compared to recorded observations) across much of the domain. These high-yield zones are enveloped by orange to Mahogany bands (64–84 % $G_y$), which themselves are bounded by cyan to yellowish-green frames corresponding to moderate yields of 24–54 %. Interstitial regions represented in Turquoise to dusky blue (>24–14 % $G_y$). Peripheral areas appear in violet to dark blue hues (>14–0 % $G_y$), with the lowest yields at the extremities of dose, monomer concentration, and temperature.

Six points of interest, A–D (experimentally validated) and B', C', D', D'' (predicted only) highlight SoDip performance near the fringes of the experimental dataset. At point A ($D_{Dose}$: 100kGy, $D_{monomer\_conc}$: 50 vol%, $D_{Temp}$: 25°C, $D_{Time}$: 40 h), $G_{y_*}$: 87.60 % vs. $G_y$: 83.14 % (Δ = 4.46 units), demonstrating slight overprediction in the highest-yield regime. Point B ($D_{Dose}$: 80kGy, $D_{monomer\_conc}$: 50 vol%, $D_{Temp}$: 25°C, $D_{Time}$: 20 h) shows $G_{y_*}$: 86.91 % vs. $G_y$: 78.29 % (Δ = 8.62 units), indicating increased deviation under mid-range conditions. Its predicted counterpart B' ($D_{Dose}$: 80kGy, $D_{monomer\_conc}$: 50 vol%, $D_{Temp}$: 43°C, $D_{Time}$: 20 h) yields $G_{y_*}$: 97.86 %, suggesting that elevating temperature would push yields well above the experimentally observed maximum. Point C ($D_{Dose}$: 30kGy, $D_{monomer\_conc}$: 30 vol%, $D_{Temp}$: 45°C, $D_{Time}$: 20 h) yields $G_{y_*}$: 55.59 % vs. $G_y$: 59.05 % (Δ = –3.46 units), whereas its prediction C' ($D_{Dose}$: 30kGy, $D_{monomer\_conc}$: 30 vol%, $D_{Temp}$: 57°C, $D_{Time}$: 20 h) gives $G_{y_*}$: 107.33 %, indicating a strong temperature sensitivity at low $D_{Dose}$ and $D_{monomer\_conc}$. Point D ($D_{Dose}$: 40kGy, $D_{monomer\_conc}$: 60 vol%, $D_{Temp}$: 25°C, $D_{Time}$: 20 h) shows $G_{y_*}$: 80.07 % vs. $G_y$: 83.18 % (Δ = –3.11 units), confirming model reliability at moderate conditions. Predicted scenarios D' ($D_{Dose}$: 50kGy, $D_{monomer\_conc}$: 60 vol%, $D_{Temp}$: 67°C, $D_{Time}$: 20 h and D'' ($D_{Dose}$: 40kGy, $D_{monomer\_conc}$: 60 vol%, $D_{Temp}$: 45°C, $D_{Time}$: 20 h) yield $G_{y_*}$: 106.61 % and 103.94 %, respectively, suggesting unexplored high-yield potential at elevated temperature or dose. Overall, the model captures the topology of experimental yields with average deviation 4.9 units and maximum deviations up to 8.6 units, accurately describing high, mid, and low-yield regions and aware that increased temperature or dose at the periphery could significantly drive yields even beyond those currently recorded.

Bhatt, A. I., Jablonka, K. M., & Smit, B. (2020). Accelerating polymer discovery with automated experimentation. Trends in Chemistry, 2(4), 335–346.

**Acknowledgment**

The results presented in this article were obtained based on a project (JPNP21012) commissioned by the New Energy and Industrial Technology Development Organization (NEDO). The authors acknowledge the computational facilities provided by the Nagoya University, Japan. The authors would like to thank Associate Prof. Clemens Heitzinger for fruitful discussion on non-parametric Bayesian analysis.

**Conflict of Interest**

All authors declare no financial or non-financial competing interests.

**Data Availability**

The datasets generated and/or analysed during the current study are not publicly available because the data were obtained from published sources and cannot be redistributed. However, processed data supporting the findings are available from the corresponding author on reasonable request.

Formation by Direct Reduction Using γ Radiation onto Silicone Surface and Their Antimicrobial Activity and Biocompatibility. *Molecules* **26**, (2021).